\documentclass[letterpaper]{article}
\usepackage{aaai23}  
\usepackage{times}  
\usepackage{helvet}  
\usepackage{courier}  
\usepackage[hyphens]{url}  
\usepackage{graphicx} 
\usepackage{natbib}  
\usepackage{caption} 
\usepackage{algorithm}
\usepackage{algorithmic}

\usepackage{newfloat}
\usepackage{listings}
\DeclareCaptionStyle{ruled}{labelfont=normalfont,labelsep=colon,strut=off} 
\floatstyle{ruled}
\newfloat{listing}{tb}{lst}{}
\floatname{listing}{Listing}
\usepackage{amsthm}
\newtheorem{remark}{Remark}

\usepackage[utf8]{inputenc} 
\usepackage{url}            
\usepackage{booktabs}       
\usepackage{amsfonts}       
\usepackage{nicefrac}       
\usepackage{microtype}      
\usepackage{mathtools}     
\usepackage{pifont}
\usepackage{mathrsfs}
\usepackage{subcaption}
\usepackage{verbatim}
\usepackage{amssymb}
\usepackage{dsfont}
\usepackage{xspace} 
\usepackage{bm}
\usepackage{multicol}
\usepackage{placeins}
\DeclareRobustCommand{\eg}{e.g.,\@\xspace}                         
\DeclareRobustCommand{\ie}{i.e.,\@\xspace}                         
\DeclareRobustCommand{\wrt}{w.r.t.\@\xspace}

\newcommand{\mathbr}[1]{\bm{\mathbf{#1}}}
\DeclareRobustCommand{\quotes}[1]{``#1''}

\newcommand{\E}{\mathop{\mathbb{E}}}
\DeclareMathOperator*{\argmin}{arg\,min}
\DeclareMathOperator*{\argmax}{arg\,max}

\newcommand{\ev}{\mathop{\mathbb{E}}}
\newcommand{\vtheta}{\mathbr{\theta}}

\newcommand{\vomega}{\mathbr{\omega}}

\DeclareRobustCommand{\algname}{WAC\@\xspace}
\DeclareRobustCommand{\algnameext}{Wasserstein Actor-Critic\@\xspace}

\title{Wasserstein Actor-Critic: Directed Exploration via Optimism for Continuous-Actions Control}
\author{
    Amarildo Likmeta, \textsuperscript{\rm 1, \rm 2}
    Matteo Sacco, \textsuperscript{\rm 2}
    Alberto Maria Metelli, \textsuperscript{\rm 2}
    Marcello Restelli, \textsuperscript{\rm 2}
}
\affiliations{
    \textsuperscript{\rm 1}University of Bologna,
    \textsuperscript{\rm 2}Politecnico di Milano\\
    amarildo.likmeta2@unibo.it, matteo3.sacco@mail.polimi.it, albertomaria.metelli@polimi.it,
    marcello.restelli@polimi.it
}
\begin{document}

\maketitle

\begin{abstract}
  Uncertainty quantification has been extensively used as a means to achieve efficient directed exploration in Reinforcement Learning (RL). However, state-of-the-art methods for continuous actions still suffer from high sample complexity requirements. Indeed, they either completely lack strategies for propagating the epistemic uncertainty throughout the updates, or they mix it with aleatoric uncertainty while learning the full return distribution (e.g., distributional RL). In this paper, we propose Wasserstein Actor-Critic (WAC), an actor-critic architecture inspired by the recent Wasserstein Q-Learning (WQL) \citep{wql}, that employs approximate Q-posteriors to represent the epistemic uncertainty and Wasserstein barycenters for uncertainty propagation across the state-action space. WAC enforces exploration in a principled way by guiding the policy learning process with the optimization of an upper bound of the Q-value estimates.
 Furthermore, we study some peculiar issues that arise when using function approximation, coupled with the uncertainty estimation, and propose a regularized loss for the uncertainty estimation. Finally, we evaluate our algorithm on standard MujoCo tasks as well as suite of continuous-actions domains, where exploration is crucial, in comparison with state-of-the-art baselines. 
\end{abstract}

\section{Introduction}
Reinforcement Learning~\citep[RL,][]{Sutton1998} is one of the most widely used frameworks for solving sequential decision-making problems.
When an agent acts in an uncertain environment, it faces the choice between \emph{exploring} with the hope of discovering more profitable behaviors or \emph{exploiting} the current information about the actions' values. This exploration-exploitation dilemma is particularly challenging in \emph{continuous-state} spaces, where \emph{function approximation} is required to generalize across states, and an accurate estimate of the uncertainty on the value estimates is not available point-wise. \emph{Continuous-action} tasks pose additional challenges since most exploration methods require the maximization of some objective (\eg upper bound of the Q-value) over the action space. While in the discrete case, this maximization can be performed by enumeration, in the continuous case it requires solving a complex optimization problem, increasing the computational demands.

\emph{Actor-Critic} (AC) methods~\citep{sac, oac, trpo} represent the current state of the art for continuous control. Despite their widespread adoption, these methods still suffer from high sample complexity. Efficient exploration strategies have been extensively studied in the literature as a means of reducing sample complexity mainly in tabular domains~\citep{ucrl,psrl,odonoghue18the,wql}.
Classical exploration strategies, like $\epsilon$-greedy or Boltzmann~\citep{Sutton1998}, inject noise around the current greedy policy to enforce exploration. Although in simple settings this is enough to guarantee convergence~\citep{szepesvary_q_learning}, this exploration strategy is not efficient in the general case. 

A common trend in the RL literature consists of endowing existing methods with some form of uncertainty quantification and using it to perform \emph{directed} exploration while focusing on the most promising regions. 
In particular, a recent extension of Soft Actor-Critic~\citep[SAC,][]{sac}, Optimistic Actor-Critic~\citep[OAC,][]{oac}, proved to improve sample efficiency over the standard SAC.
Indeed, uncertainty quantification is a fundamental step to define efficient exploration strategies. Exploration strategies, coming from the Multi-Armed Bandit~\citep[MAB,][]{mab_tor} literature, have been extended for the RL settings, starting from tabular domains~\citep{ucrl, psrl, wql}, with theoretical guarantees on the sample complexity and/or regret. In turn, they have been extended to the Deep Reinforcement Learning (DRL) settings too, but the guarantees no longer hold up. 
Ensemble methods allow quantifying the uncertainty but do not \emph{propagate} it across the state action-space when performing the critic updates. Uncertainty propagation is a fundamental tool of any principled uncertainty estimation approach since most AC methods rely on bootstrapping when updating the critics. This results in Q-value estimates that also incorporate uncertainty about the bootstrapped values. Distributional RL~\citep{odonoghue18the} allows for uncertainty propagation but considers only aleatoric uncertainty, being aimed at estimating the full return distribution. 

In this paper, we address the problem of uncertainty estimation and propagation in the context of continuous-action RL. Starting from the methodology introduced in WQL~\citep{wql}, we devise a novel actor-critic algorithm, \emph{Wasserstein Actor-Critic} (\algname), which employs Q-posteriors both to quantify uncertainty on the critic estimates to drive exploration, as well as a tool to propagate it across the state-action space~(Section~3). 
Furthermore, we consider some practical problems that arise while quantifying uncertainty by means of Q-posteriors coped with function approximators, especially neural networks. To this end, we propose a regularization approach for the uncertainty networks (Section~4).
After reviewing the literature (Section~5), we present a thorough experimental evaluation over some simple 1D navigation domains, as well as some MujoCo~\citep{mujoco_todorov} tasks designed for exploration to assess the effect of uncertainty estimation and propagation on exploration and sample complexity (Section~6). 

\section{Preliminaries}
\label{sec:preliminaries}

\paragraph{Markov Decision Processes}
We consider infinite-horizon discounted Markov Decision Processes \citep[MDP,][]{puterman2014markov}. An MDP is a 5-tuple $\mathcal{M}=(\mathcal{S},\mathcal{A},\mathcal{P},\mathcal{R},\gamma)$, defined by the state space $\mathcal{S}$, the action space $\mathcal{A}$, a transition kernel $\mathcal{P}: \mathcal{S} \times \mathcal{A} \rightarrow \Delta(\mathcal{S})$, a reward $\mathcal{R}: \mathcal{S} \times \mathcal{A} \rightarrow \Delta(\mathbb{R})$ and the discount factor $\gamma \in {[0,1)}$.\footnote{$\Delta(\mathcal{X})$ denotes the set of probability distributions over $\mathcal{X}$.} Let $r: \mathcal{S} \times \mathcal{A} \rightarrow \mathbb{R}$ be the expectation of the reward $\mathcal{R}$ that we assume bounded in $[r_{\min},r_{\max}]$. The behavior of an agent is described by a policy $\pi: \mathcal{S} \rightarrow \Delta(\mathcal{A})$.
The performance of a policy $\pi$ is measured by its action-value function defined as $Q^\pi(s,a) = \ev\left[ \sum_{t=0}^{\infty} \gamma^t r_t \right | s_0=s, a_0=a]$, where we fix the first action and follow policy $\pi$ for the next steps. The value functions satisfy the Bellman equations, $Q^{\pi}(s,a) = r(s,a) + \gamma \ev \left[ Q^{\pi}(s',a') \right] $ for every $ (s,a) \in \mathcal{S} \times \mathcal{A}$. The optimal action-value function $Q^*$ is the maximum, over all policies, of the $Q$ function, for all state-action pairs,  $Q^*(s,a) = \sup_{\pi} \{Q^{\pi}(s,a)\}$. $Q^*$ satisfies the Bellman optimality equation $Q^*(s,a) = r(s,a) + \gamma \ev \left[ \sup_{a' \in \mathcal{A}} \{Q^*(s',a')\} \right]$. The Bellman equations form the basis of Temporal Difference (TD) learning, which updates the estimation of the $Q$ function in the current state using estimates of the next-states $Q$ function. The goal of learning algorithms in this setting is to find the optimal policy $\pi^*$, which is defined as the policy that acts greedily \wrt $Q^*$, $\pi^*(\cdot|s) \in \Delta\left(\argmax_{a \in \mathcal{A}} \{Q^*(s,a)\}  \right), \forall s \in \mathcal{S}$.

\paragraph{Actor-Critic Methods}
AC methods maintain a parameterized value-function $Q_{\vomega}$ (\emph{critic}) to estimate the value of the current (or a given target) policy, and a parameterized policy $\pi_{\vtheta}$ (\emph{actor}), trained through gradient descent. In particular, 
SAC, employs an \emph{entropy-regularized} architecture, \ie the agents optimize a modified objective regularized with the entropy of the policy favoring stochastic policies over deterministic ones, shown in Equation~\ref{eq:sac_obj}. Specifically, it maintains two parameterized action-value functions $\{{Q}_{\vomega_1}, {Q}_{\vomega_2}\}$ to estimate the entropy-regularized value function of policy $\pi_{\vtheta}$. They are trained on the same samples and differ only on the initialization of $\vomega_1$ and $\vomega_2$. The actor optimizes a \quotes{lower bound} of the action-value function, ${Q}_{LB}(s,a) = \min\{{Q}_{\vomega_1}(s,a), {Q}_{\vomega_2}(s,a)\}$. To update the critic, given a sample $(s,a, r, s')$, SAC uses the SARSA~\citep{Sutton1998} update rule, ${Q}_{\{\vomega_1,\vomega_2\}}(s,a) \leftarrow r + \gamma {Q}_{LB}(s', a')$, where $a' \sim \pi_{\vtheta}(s')$. Specifically, SAC maintains experience collected with previous policies $\pi_{\vtheta}$ in a \emph{replay buffer} $\mathcal{D}$~\citep{Sutton1998}. The critic is trained to minimize the (entropy regularized) Bellman error over this replay buffer, as follows:
\begin{equation}
\begin{split}
    J_{\text{C}}(\{\vomega_1,\vomega_2\}) = \ev_{s,a,r,s' \sim \mathcal{D}} \Big[ & (Q_{\{\vomega_1,\vomega_2\}}(s,a) -  \\ & (r + \gamma \widetilde{Q}(s',a')))^2 \Big],
\end{split}
\end{equation}
where $\widetilde{Q}(s,a) = \overline{Q}_{LB}(s,a) - \alpha \log \pi_{\vtheta}(s',a')$, $\overline{Q}_{LB}(s,a) = \min\{{Q}_{\overline{\vomega}_1}(s,a), {Q}_{\overline{\vomega}_2}(s,a)\}$ is the lower bound of the Q-values given by two target networks which are updated slowly to improve stability~\citep{mnih2015humanlevel}, $a' \sim \pi_{\vtheta}(s')$, and $\alpha>0$ specifies the level of entropy regularization. The actor network is trained to optimize an entropy-regularized objective. Since the target $Q$-function is a parameterized function approximator, the policy can directly follow the gradient of the critic:
\begin{equation}
    \label{eq:sac_obj}
    J_A(\vtheta) = \ev_{\substack{s_t \sim \mathcal{D} \\ a_t \sim \pi_{\vtheta}(s_t)}} \left [ \log \pi_{\vtheta}(s_t, a_t) - {Q}_{LB}(s_t, a_t) \right].
\end{equation}

\paragraph{Wasserstein TD-Learning}
Bayesian approaches to RL~\citep{dearden1998bayesian,wql} maintain, for each state-action pair $(s,a) \in \mathcal{S} \times \mathcal{A}$, a probability distribution $\mathcal{Q}(s,a)$, called a \emph{Q-posterior}, used to represent the \emph{epistemic} uncertainty over the value estimates. In practice, $\mathcal{Q}$ is an approximate distribution in a class $\mathscr{Q}$ (\eg Gaussians). 
Using the concept of barycenter, we can also propagate the uncertainty of the value function estimates across the state-action space.
As in~\cite{wql}, we employ the notion of barycenter defined in terms of Wasserstein~\citep{villani2008optimal} divergence since the variance of our Q-posteriors vanishes as the number of samples grows to infinity, and the Wasserstein divergence allows computing the distance between distributions with disjoint support. Given two probability distributions, $\mu$ and $\nu$, the $L_p$-Wasserstein distance between $\mu$ and $\nu$ is defined as: $W_p(\mu,\nu) = ( \inf_{\rho \in \Gamma(\mu,\nu)} \E_{X,Y \sim \rho} [ d(X,Y)^p] )^{1/p}$, where $\Gamma(\mu,\nu)$ is the set of all joint distributions with marginals $\mu$ and $\nu$, and $d$ is a metric (\ie we use the $L_2$-norm). Given a class of probability distributions $\mathscr{N}$, a set of probability distributions $\{\mu_i\}_{i=1}^n$, $\mu_i \in \mathscr{N}$ and a set of weights $\{\xi_i\}_{i=1}^n$, $\sum_{i=1}^n \xi_i = 1$ and $\xi_i \ge 0$, the $L_{2}$-Wasserstein barycenter is defined as~\citep{agueh2011barycenters}: $\overline{\mu} \in \argmin_{\mu \in \mathscr{N}} \{ \sum_{i=1}^n \xi_i W_2(\mu, \mu_i)^2 \}$.
From this, having observed a transition $(s,a,r,s')$, the \emph{Wasserstein Temporal Difference}~\citep[WTD,][]{wql} update rule is defined via the computation of the barycenter of the current Q-posterior and the \emph{TD- target posterior}, defined as $\mathcal{T}_t = r + \gamma \mathcal{Q}_t(s', a')$:
\begin{equation}
\label{eq:wtd}
    \begin{split}
	\mathcal{Q}_{t+1}(s,a) \in  \argmin_{\mathcal{Q} \in \mathscr{Q}} \Big\{ & (1-\alpha_t) W_2 \left(\mathcal{Q}, \mathcal{Q}_t(s,a) \right)^2 + \\
	& \alpha_t W_2 \left(\mathcal{Q}, \mathcal{T}_t \right)^2 \Big\},
	\end{split}
\end{equation}
where $\alpha_t$ is the learning rate, and $a'$ is the action taken in the next step. Depending on the policy used to select action $a'$, the update can be either on-policy or off-policy. The presence of $\gamma$ in the definition of $\mathcal{T}_t$ shrinks the posteriors, vanishing the uncertainty when the number of samples grows to infinity. When the Q-posteriors become point estimates, the update rule reduces to the classic TD update rule. 

\section{Wasserstein Actor-Critic}
\label{sec:wac}
 \begin{algorithm}[tb]
 \small\captionof{algorithm}{\algnameext.}
 \label{alg:wac}
 \begin{algorithmic}
 \STATE {\bfseries Input:}  critic parameters $\vomega_1,\vomega_2$, policy parameters $\vtheta, \vtheta^T$
  \STATE Initialize $\mathcal{Q}_{\{1,2\}}(s,a)$ with the prior $\mathcal{Q}_0$
    \STATE Initialize replay buffer $\mathcal{D} \leftarrow \emptyset$
  \FOR{$\text{epoch} = 1,2, ...$}
        \FOR{$t = 1,2,...$}
          \STATE Take action $a_t \sim \pi_{\vtheta}(\cdot|s_t)$
          \STATE Observe $s_{t+1}$ and $r_{t+1}$
          \STATE $\mathcal{D} \leftarrow \mathcal{D} \cup \{ (s_t, a_t, r_{t+1}, s_{t+1})\}$
        \ENDFOR
        \STATE $\sigma_{\text{old}}^{\{1,2\}} \leftarrow \sigma_{\vomega_{\{1,2\}}}$
        \FOR{$\text{iteration} = 1,2, ...$}
          \STATE Update critic weights $\vomega_{\{1,2\}}$ using Equation~\eqref{eq:regularized_critic_loss}
          \STATE Update actor (resp. target) weights $\vtheta$ (resp. $\vtheta^T$) using Equation~\eqref{eq:actor_obj} (resp. Equation~\eqref{eq:actor_obj_target})
        \ENDFOR
  \ENDFOR
  \end{algorithmic}
\end{algorithm}
In this section, we introduce Wasserstein Actor-Critic (\algname), which extends WQL to handle environments with continuous-action spaces. We present the algorithm, define the update rules, and a regularization for the uncertainty estimates. 

\paragraph{Distributional Critic}
For each state-action pair $(s,a) \in \mathcal{S} \times \mathcal{A}$, we maintain an approximate distribution $Q(s,a) \sim \mathcal{Q}(s,a)$, to model the uncertainty estimate on the value function. While these distributions will generally depend on the aleatoric uncertainty of the environment (state transition and reward), our updates will vanish the variance as we collect samples. This represents our main difference w.r.t. Distributional RL~\cite{distributionalRL}, as we do not require learning the whole return distribution, while still propagating uncertainty across the state-action space.
More specifically, given a replay buffer of past behavior $\mathcal{D}$, our critic minimizes the $L_2$-Wasserstein distance between the Q-posterior $ \mathcal{Q}_{\vomega}$ and the target posterior $r + \gamma \mathcal{Q}_{\overline\vomega}$, defined through the target parameters $\overline\vomega$ and target policy $\pi_{\vtheta^T}$:
\begin{equation}
    \label{eq:critic_obj}
    \begin{aligned}
    J_C({\vomega})  =  \E_{s,a,s',r \sim \mathcal{D}} & \Big[ W_2 \Big( \mathcal{Q}_{\vomega}(s,a), \\ 
    & \qquad r + \gamma \mathcal{Q}_{\overline\vomega}(s',\pi_{\vtheta^T}(s') \Big)^2 \Big].
    \end{aligned}
\end{equation}
Different flavors of the algorithm can be proposed, based on the combination of: (i) distribution classes $\mathscr{Q}$, (ii) behavioral policy $\pi_{\vtheta}$, and (iii) target policy $\pi_{{\vtheta}^T}$. 
We focus on optimistic exploration, that requires optimizing upper bounds. Moreover, although other distribution classes, like particle-models, could be employed, we limit our discussion to Gaussian posteriors, as their parametrization allows for direct control over the distribution variance.

Similar to WQL, we maintain a parameterized distributional critic using a function approximator (e.g., neural network) that outputs the parameters of the distribution. For the Gaussian case, ${Q}(s,a) \sim \mathcal{N}(\mu_{\vomega}(s,a), \sigma_{\vomega}(s,a))$, the Wasserstein distance has a closed form, and the critic objective becomes:
\begin{align}
    J_C({\vomega}) = & \E_{s,a,s',r \sim \mathcal{D}} \Big[\big(  \mu_{\vomega}(s,a) - (r + \gamma \widetilde{\mu}_{\overline\vomega}(s',\pi_{{\vtheta}^T}(s'))) \big)^2 \notag\\ 
    & \quad + \big(\sigma_{\vomega}(s,a) - \gamma \sigma_{\overline\vomega}(s',\pi_{{\vtheta}^T}(s')) \big)^2 \Big],\label{eq:gaussian_critic_obj}
\end{align}
where $\widetilde{\mu}_{\overline\vomega}(s,a) = \mu_{\overline\vomega}(s,a) - \alpha \log \pi_{{\vtheta}^T}(s,a)$. 
In practice, $\mu_{\vomega}$ and $\sigma_{\vomega}$ can use either a shared network architecture or two different networks. We initialize the posterior networks using the bias of the last layer of the network. If the reward function is limited in the interval $[r_{\min},r_{\max}]$, the Q values will be in the range $[q_{\min}, q_{\max}]$ with $q_{\min} = r_{\min}/(1 - \gamma)$ and  $q_{\max} = r_{\max}/({1 - \gamma})$. We therefore initialize the uncertainty networks to $\sigma_0 = (q_{\max} - q_{\min})/\sqrt{12}$, i.e. variance of the Gaussian minimizing the KL divergence with the uniform distribution in $[q_{\min}, q_{\max}]$~\citep{wql}.

\paragraph{Actor}
The actor in WAC is updated by optimizing an \emph{upper bound} $U^{\delta}_{\vomega}$ of the estimated Q-value, which we can efficiently compute using Gaussian posterior: $U^{\delta}_{\vomega}(s,a) = \mu_{\vomega}(s,a) + \sigma_{\vomega}(s,a) \Phi^{-1}(\delta)$, where $\Phi^{-1}$ is the quantile function of the standard normal and $\delta \in (0, 1)$. When actions are finite, no actor is needed, as we can compute the maximum by enumeration. However, in the continuous-action case, we need an actor that follows $U^{\delta}_{\vomega}(s,a)$, which is differentiable in $\vomega$, leading to the minimization of the objective:
\begin{equation}
\label{eq:actor_obj}
        J_A(\vtheta) = \ev_{\substack{s_t \sim \mathcal{D} \\ a_t \sim \pi_{\vtheta}(s_t)}} \left [ \log \pi_{\vtheta}(s_t, a_t) - U^{\delta}_{\vomega}(s_t, a_t) \right],
\end{equation}
where $\vtheta$ are the parameters of the behavioral policy. 

\paragraph{Target Policy}
We propose two alternatives for the target policy $\pi_{{\vtheta}^T}$, corresponding to different estimators for the target posterior $
\mathcal{T}_t$. First, we can use the same policy we use for exploration, i.e., $\vtheta={\vtheta}^T$, like SAC. This has the advantage of not requiring a second parameterized policy. We call this version \emph{Optimistic Estimator-WAC} (OE-WAC), which represents an on-policy algorithm. Alternatively, we can use a \emph{greedy} policy that optimizes the expected value of the Q-posteriors (the mean critic  $\mu_{\vomega}(s,a)$ in the Gaussian case). In this case, the target policy minimizes:
\begin{equation}
    \label{eq:actor_obj_target}
        J_T(\vtheta_T) = \ev_{\substack{s_t \sim \mathcal{D} \\ a_t \sim \pi_{\vtheta^T}(s_t)}} \left [ \log \pi_{\vtheta^T}(s_t, a_t) - \mu_{\vomega}(s_t, a_t) \right].
\end{equation}
We call this version \emph{Mean Estimator-WAC} (ME-WAC). The best version to use between the two is task-dependent. Generally, OE-WAC is more suitable for environments that require large exploration, whereas ME-WAC is more suitable for simpler environments where OE-WAC might over-explore and might suffer from some instability. 

\begin{remark}[What is our Critic Estimating?]
We underline that our distributional critic maintains uncertainty about the $Q^*$ and not about the Q-function of the current policy $Q^{\pi_{\vtheta}}$. Indeed, the method starts with an initial high-uncertainty estimate and updates it as we collect samples from the environment. In the tabular case, it can be proven that these upper-bounds on the value of $Q^*$ are valid with high probability for every timestep $t$, under some conditions on the learning rate $\alpha_t$~\cite{wql}. When extending the method to DeepRL, these guarantees are no longer valid, since the uncertainty estimates are outputs of general function approximators and local updates are no longer possible.
\end{remark}

\section{Regularized Uncertainty Estimation}\label{regularized_uncertainty}
\begin{figure}
\centering
	\includegraphics[width=\linewidth]{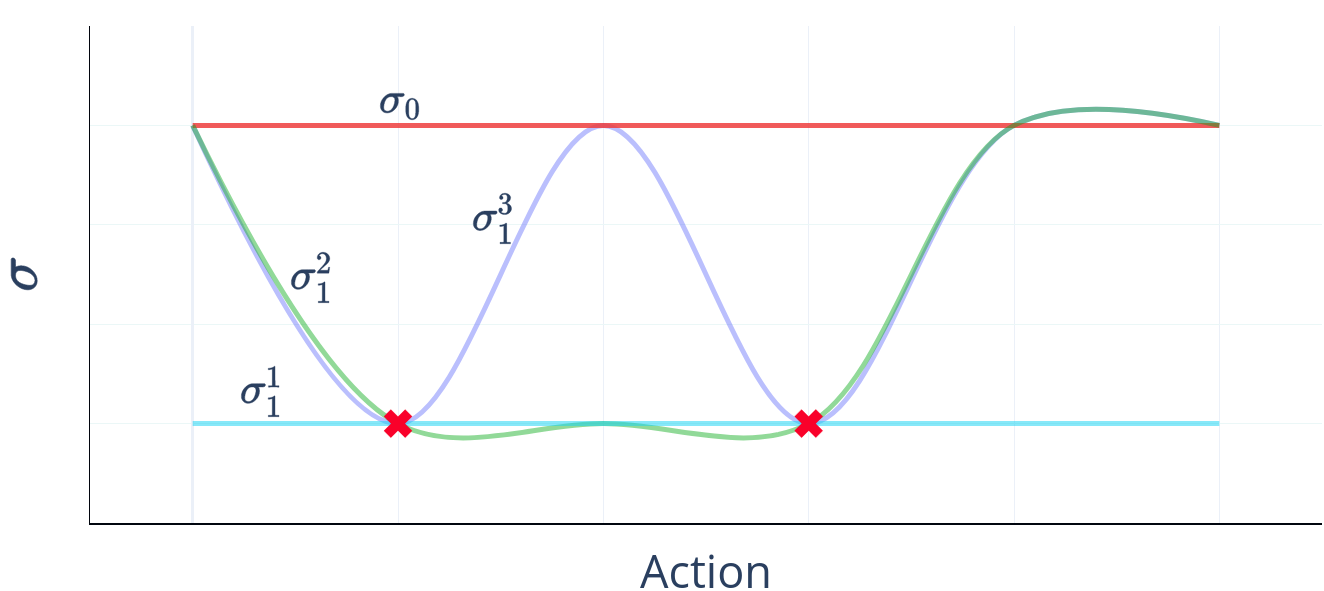}
\caption{Example of uncertainty estimates. $\sigma_0$ shows the initial constant high value.}\label{fig:reg_uncertainty}
\end{figure}
Our Q-posteriors are initialized to high uncertainty at the beginning of the learning process. Since they represent epistemic uncertainty, their variance will shrink as we observe more samples. This is apparent in Equation~\eqref{eq:gaussian_critic_obj}, where the targets $\sigma_{\vomega}$ are multiplied with $\gamma$. In tabular settings, the updates are localized, i.e., they affect a single state-action pair, without interfering with the others. However, when function approximators are involved, generalizing uncertainty in an uncontrolled way might cause non-visited areas of the state-action space to take low uncertainty values, which might be undesired.\footnote{This generalization phenomenon happens for the mean too, but, as visible in Equation~\eqref{eq:gaussian_critic_obj}, is particularly critical for the variance that gets updated with the next-state-action variance scaled by $\gamma <1$.}
Consider the example in Figure~\ref{fig:reg_uncertainty} showing the uncertainty estimate as a function of the action, in a fixed state $s$. Starting from an initial high constant estimate of $\sigma_0$, at the beginning of the learning process, we will observe samples like the red crosses in the figure, \ie with lower uncertainty since it gets shrunk with $\gamma$. Among all the possible fitting lines, we would prefer an estimate like $\sigma_1^{3}$, which keeps high uncertainty in unseen regions, and would like to avoid failures like $\sigma_1^{1}$. This requires controlling the \quotes{smoothness} properties of the approximator.
To avoid the additional computational burden, we propose a simple scheme based on \emph{synthetic} samples. 
Specifically, we periodically save the weights of the uncertainty network $\sigma_{\text{old}}$ and use it as the target for state-action pairs drawn \emph{uniformly} from the state-action space. More formally, our distributional critic minimizes:
\begin{equation}
    \label{eq:regularized_critic_loss}
    \begin{split}
    & J'_C(\vomega_{\{1,2\}})   =  J_C(\vomega_{\{1,2\}}) + \\
    & \qquad \lambda \E_{s,a \sim \mathcal{U}(\mathcal{S} \times \mathcal{A})} \left[ \left(\sigma_{\vomega_{\{1,2\}}}(s,a) - \sigma_{\text{old}}(s,a \right)^2 \right],
    \end{split}
\end{equation}
where $J_C(\vomega_{\{1,2\}})$ is defined in Equation~\eqref{eq:gaussian_critic_obj} and $\lambda\ge 0 $ defines the relative weight of the regularization. Furthermore, in practice we add a second parameter, $\rho \in [0,1]$ which represents the fraction of \emph{fake} samples (w.r.t. the samples used for $J_C(\vomega_{\{1,2\}})$) drawn for regularization. Specifically, if we estimate $J_C(\vomega_{\{1,2\}})$ using $N$ samples from replay buffer $\mathcal{D}$, we will estimate the expectation in Equation~\eqref{eq:regularized_critic_loss} with $M=\rho N$ samples from $\mathcal{U}(\mathcal{S} \times \mathcal{A})$. 
Algorithm~\ref{alg:wac} reports the pseudocode of WAC, embedding the regularized uncertainty estimation.

\begin{figure*}
        \centering
        \begin{subfigure}[b]{=0.33\textwidth}
            \centering
            \includegraphics[width=\textwidth]{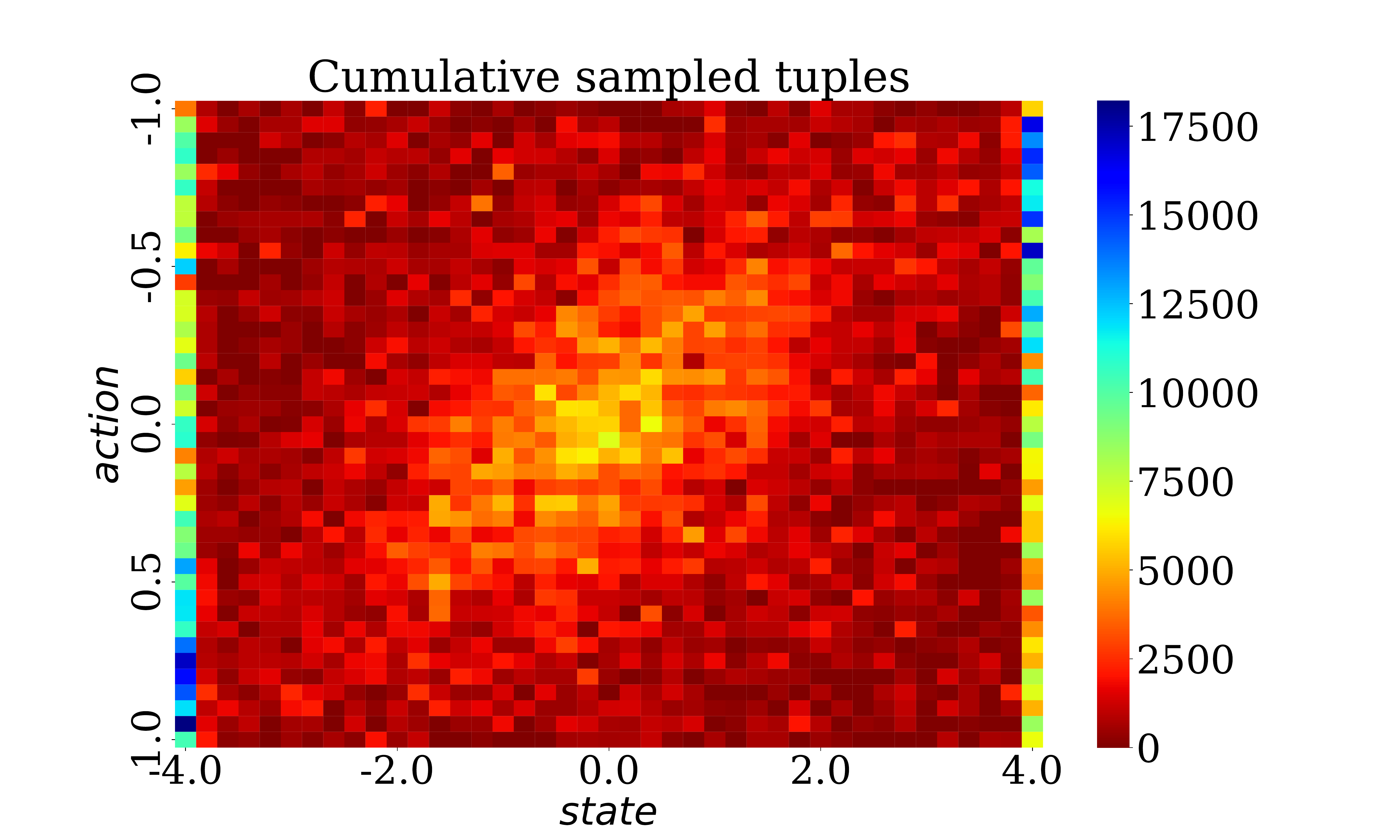}
            \caption{{State-action density}}%
            \label{fig:histogram}
        \end{subfigure}
        \hfill
        \begin{subfigure}[b]{=0.33\textwidth}
            \centering
            \includegraphics[width=\textwidth]{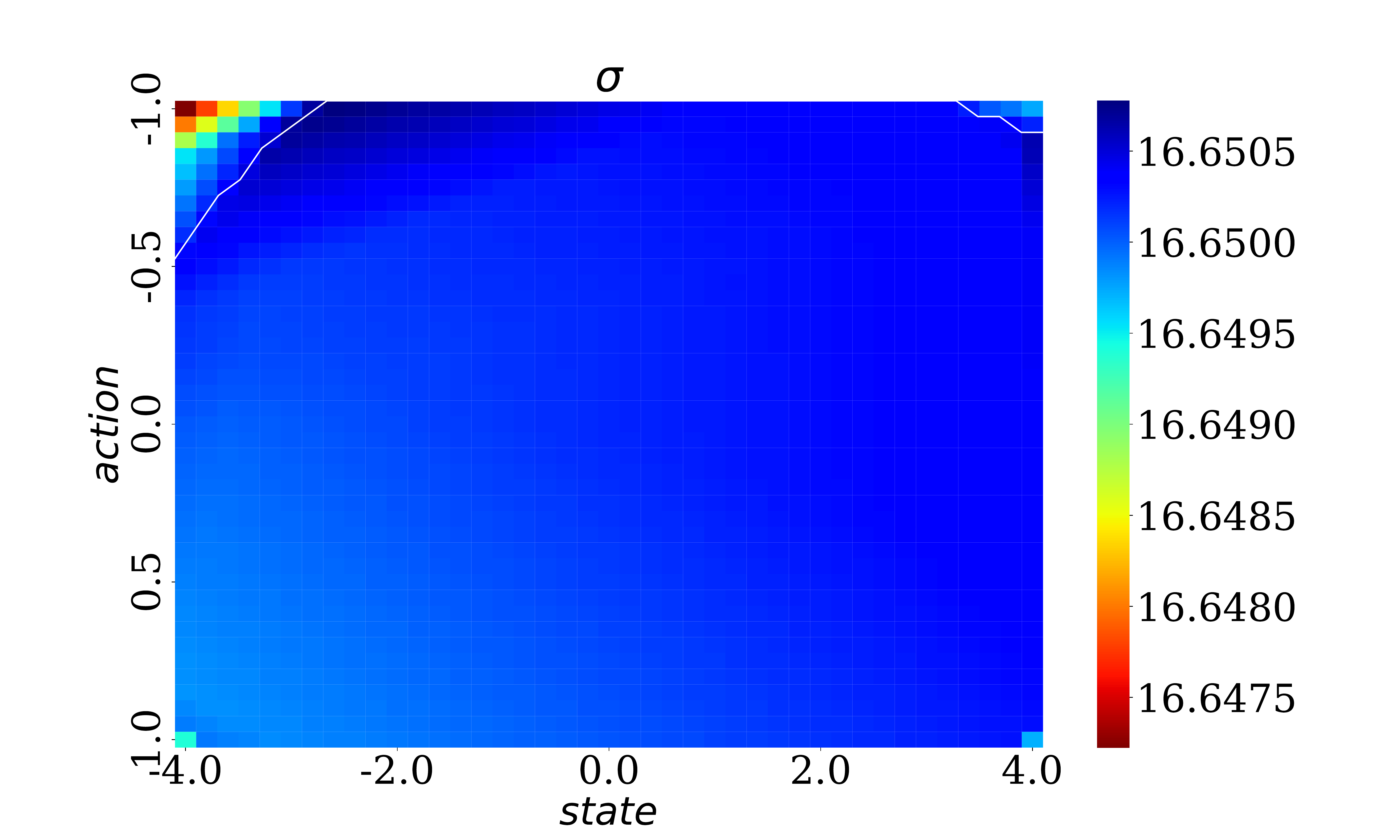}
            \caption{{No uncertainty regularization}}%
            \label{fig:histogram_std_no_regularization}
        \end{subfigure}
        \hfill
        \begin{subfigure}[b]{=0.33\textwidth}
            \centering
            \includegraphics[width=\textwidth]{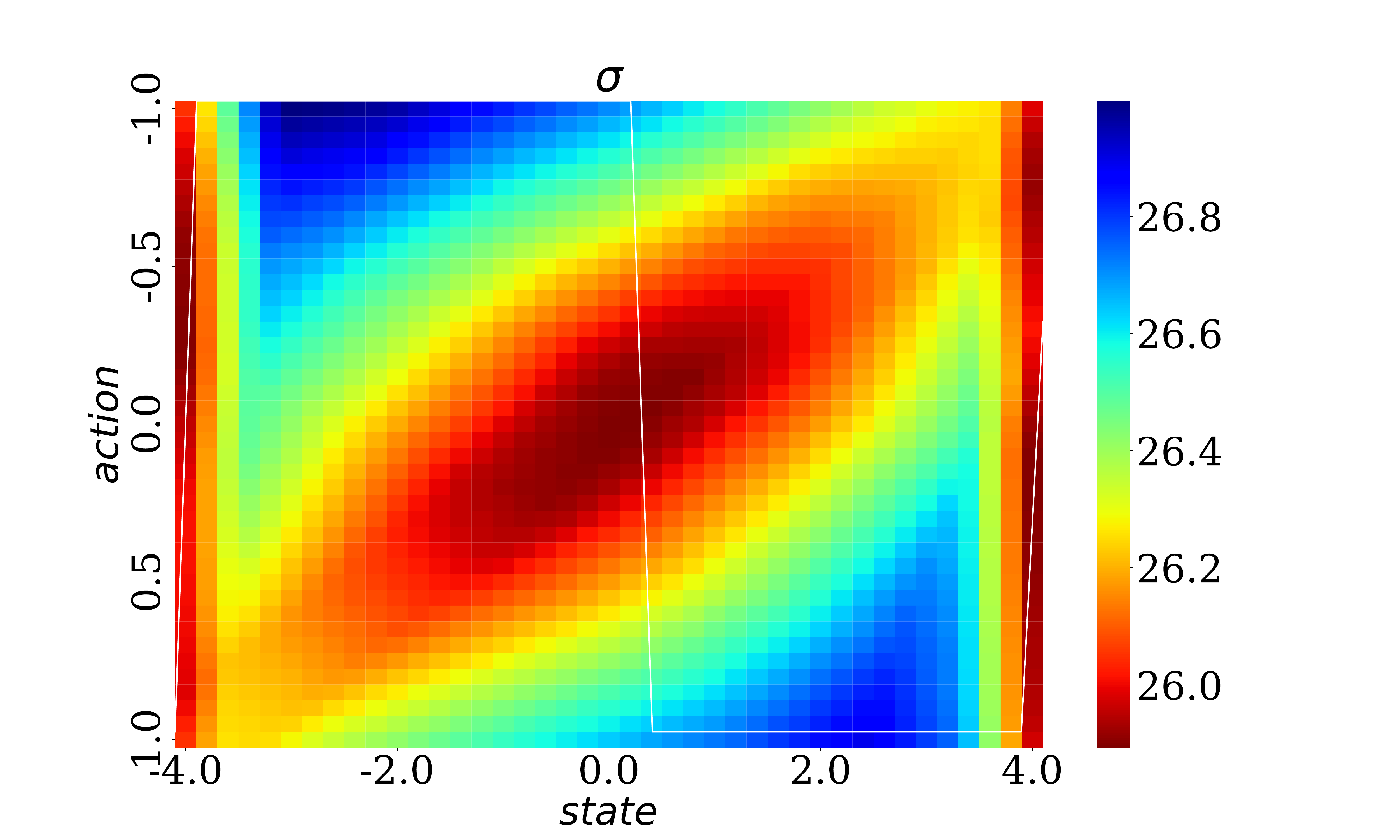}
            \caption{{With uncertainty regularization}}%
            \label{fig:histogram_std_regularized}
        \end{subfigure}
        \caption{Comparison on the uncertainty estimates after training with and without uncertainty regularization in the LQG illustrative example (action on the $y$ axis and state on the $x$ axis).}
        \label{fig:fake_samples_effect}
\end{figure*}

To investigate the effectiveness of the regularized uncertainty loss, on an illustrative example, we trained two different agents, in a one-dimensional Linear Quadratic Regulator~\citep[LQG,][]{dorato2000linear}. This task has a one-dimensional state and action spaces, which allows us to visualize the uncertainty estimates.
Figure~\ref{fig:fake_samples_effect} shows the resulting uncertainty estimates. 
On the left, we show the empirical state-action visitation distribution. The agent starts in one of the borders of the state space and has to reach the center in a few steps while calibrating the actions. This is apparent in the histogram, with the highest densities in the borders and the center. We consider it desirable to obtain uncertainty estimates that mirror these state-action densities, as the epistemic uncertainty is inversely proportional to the state-action visitation. While in both cases, the state-action densities are similar, the uncertainty estimates are completely different. In Figure~\ref{fig:histogram_std_no_regularization}, we see that without regularization, the critic completely fails to represent the uncertainty. In Figure~\ref{fig:histogram_std_regularized}, we can see that the regularized uncertainty critic, almost perfectly matches the state-action densities. In Section~6 and Appendix B, 
we show a more thorough investigation of the effect of the regularized uncertainty loss.

\begin{remark}[Where does WAC differ from WTD?]
While WAC is a direct extension of Wasserstein TD-learning to the actor-critic architecture, it does come with some modifications (mostly to regularize the learning process). Indeed, the entropy setting we employ in the critic and actor fit is not a part of the base framework. Wasserstein TD-learning, in it's pure form, would employ deterministic policies. In practice, we observed that this, coupled with optimistic exploration, caused instability in the learning process. Adding entropy regularization improves stability while not decreasing significantly the exploration capabilities of the method. Moreover, the regularization term in Equation~\eqref{eq:regularized_critic_loss} is added due to the use of general function approximators for generalizing uncertainty, and was not found in the main Wasserstein TD-framework.
We use the regularization term as a way to control the “smoothness” of the function approximator. The synthetic samples do not necessarily need to be “valid” state-action pairs. They need to be in the input space of the approximator. Indeed, in our implementation we simply uniformly sample in  ${\left[-1, 1 \right]}^{nS+nA}$ (where $nS$ and $nA$ are the dimension of the state and action space, each normalized in $\left[-1, 1\right]$). 
\end{remark}
    
\section{Related Works}
\label{sec:ralated_works}
There exists a large body of literature studying efficient exploration techniques in RL. In the tabular settings, \emph{provably efficient} methods have been devised, both in the model-based~\citep{uclr2, psrl} and model-free~\citep{strehl2006pac,chi2018is} settings. These methods cannot be easily extended to the Deep RL setting, or when extensions are proposed, they lose their theoretical guarantees. In this section, we focus on tractable exploration methods proposed for Deep RL for continuous action spaces. Two main exploration frameworks exist: \emph{uncertainty-based} methods and \emph{intrinsic motivation} methods~\citep{vime, zhang2021made, mutti2021task}. For space reasons, we will focus our discussion on uncertainty-based methods.

Classical value-based methods (including ACs) maintain a point estimate of the value functions for each state (or state-action pairs).
Exploration policies, like $\epsilon$-greedy or Boltzmann~\citep{Sutton1998}, add noise around the greedy action derived from these point estimates. These methods are not efficient, mainly because the exploration is not \emph{directed} towards unvisited regions of the state space. The entropy regularization of SAC is a form of undirected exploration too, as the policies are trained to sacrifice some returns to preserve stochastic behavior. In recent years, several methods that move away from point estimates have been proposed. Ensemble methods~\citep{chen2021randomized, wang_adaptive_ensemble} implicitly model the epistemic uncertainty of the Q-value estimates by maintaining multiple Q-function approximators. OAC~\citep{oac} explicitly models the uncertainty on the value estimates by computing the variance of two critics, and it uses it to compute an exploration policy that optimizes an upper bound of the $Q$-values. Unfortunately, this uncertainty estimate is just heuristic and only stems from the disagreement between the two $Q$-networks with different initialization. Indeed, the networks are also trained with the same samples, and same target $Q$-values, so any disagreement is purely due to the random initialization only. Recently, SUNRISE~\citep{sunrise} proposes a framework to unify ensemble methods for epistemic uncertainty estimation and shows considerable performance improvements in discrete and continuous action spaces. Distributional RL, on the other hand, models the \emph{aleatoric} uncertainty, as its goal is to estimate the whole return distribution. First proposed for problems with a discrete action space~\citep{distributionalRL, qr_rl, pmlr-v97-mavrin19a}, it has been successfully extended also to the AC setting in TOP~\citep{tactical_optimism_pessimism_2021}. TOP models both aleatoric and epistemic uncertainty and adapts the level of optimism/pessimism by means of a MAB approach.
While TOP deals with uncertainty propagation, it mixes the epistemic and aleatoric uncertainty while estimating the return distribution. 

\section{Experiments}
\label{sec:experiments}
In this section, we present the empirical evaluation of \algname in various continuous control domains. We start from simple 1D-navigation,
where we can better visualize the effects of the Q-posteriors in the learning and exploration process. In Appendix B, we show an evaluation on several standard MujoCo tasks, which show that this suite of environments does not pose significant exploration challenges. Hence, we focus our evaluation of WAC on a set of MujoCo tasks specifically designed for exploration. 

\paragraph{1D Navigation}
To measure the effect of uncertainty estimation on exploration we keep track of the cumulative \emph{coverage} of the state-action space, \ie the portion of the total volume visited with relative frequency larger than $\epsilon > 0$. 
We consider a one-dimensional LQG, an environment with no particular exploration challenges, and a more challenging continuous-action version of the Riverswim~\citep{strehl2008an}, where long sequences of rewardless actions are needed to reach high reward states. A full description of the environments is reported in Appendix A.
\begin{figure*}[htp]
        \centering
        \begin{subfigure}[b]{0.49\textwidth}
            \centering
            \includegraphics[width=0.49\textwidth]{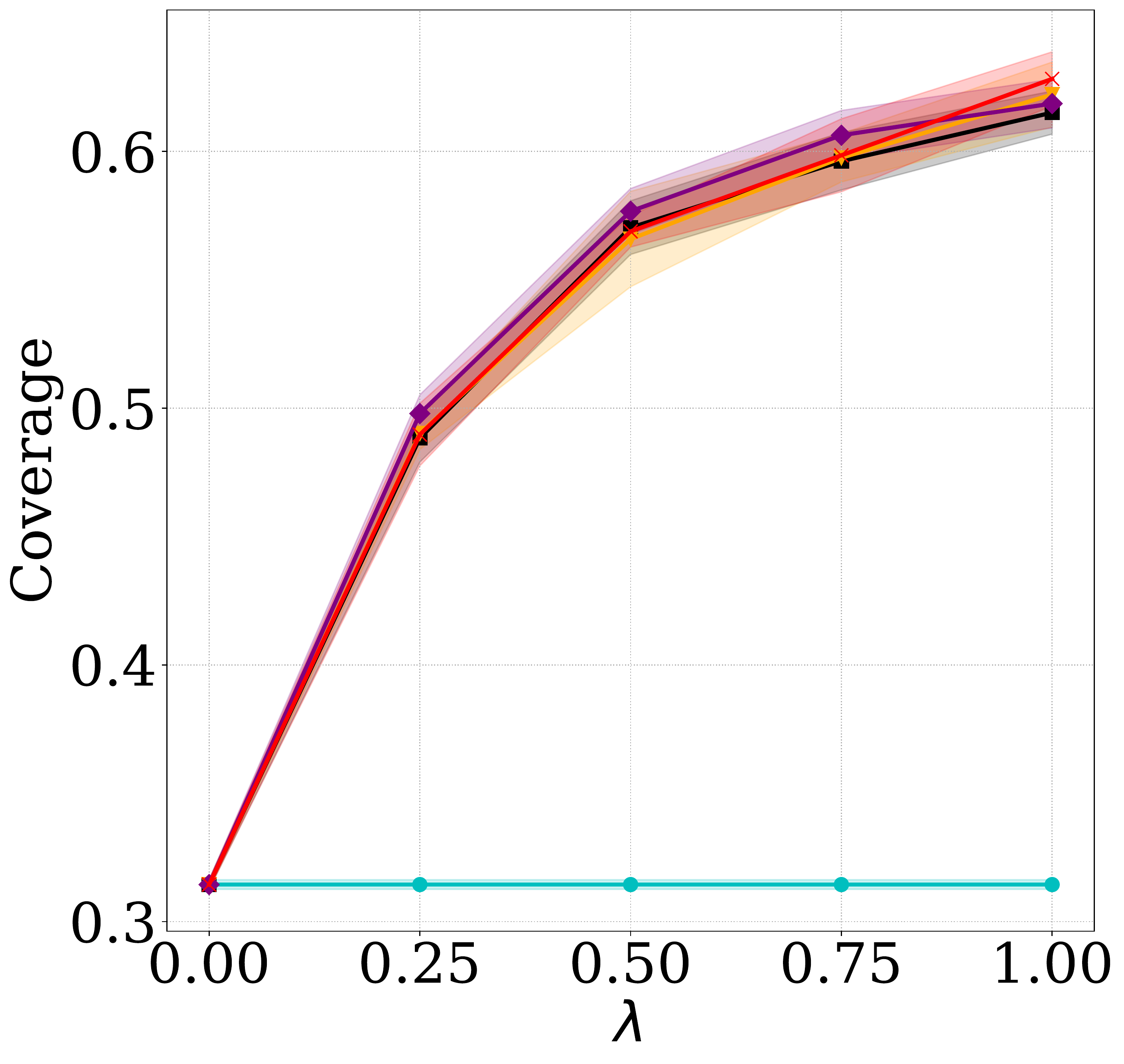}
            \hfill
            \includegraphics[width=0.49\textwidth]{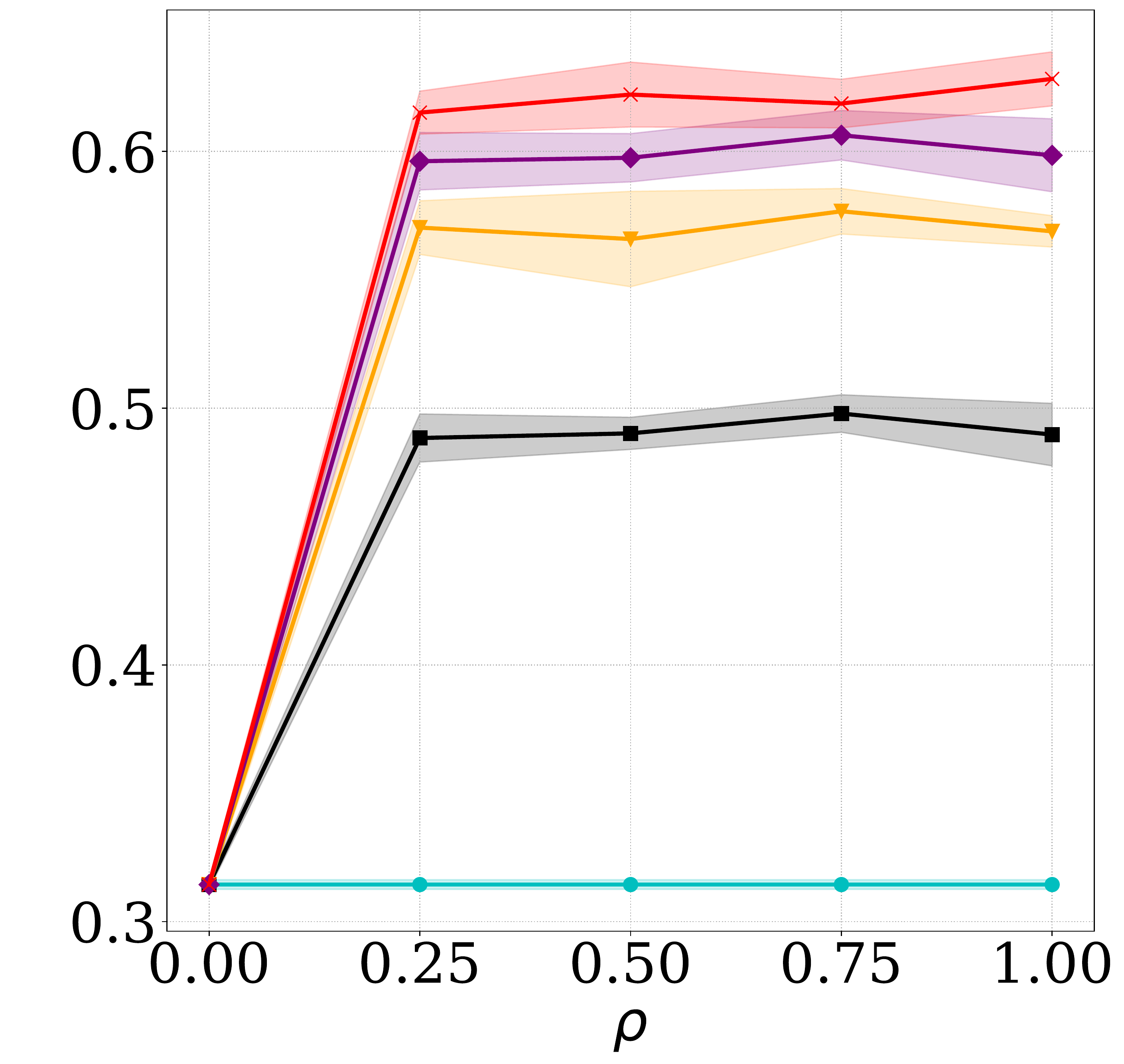}
            \caption{Coverage in LQG.}

            \label{fig:covergae_lqg}
        \end{subfigure}
        \hfill
        \begin{subfigure}[b]{0.49\textwidth}   
            \centering 
            \includegraphics[width=0.49\textwidth]{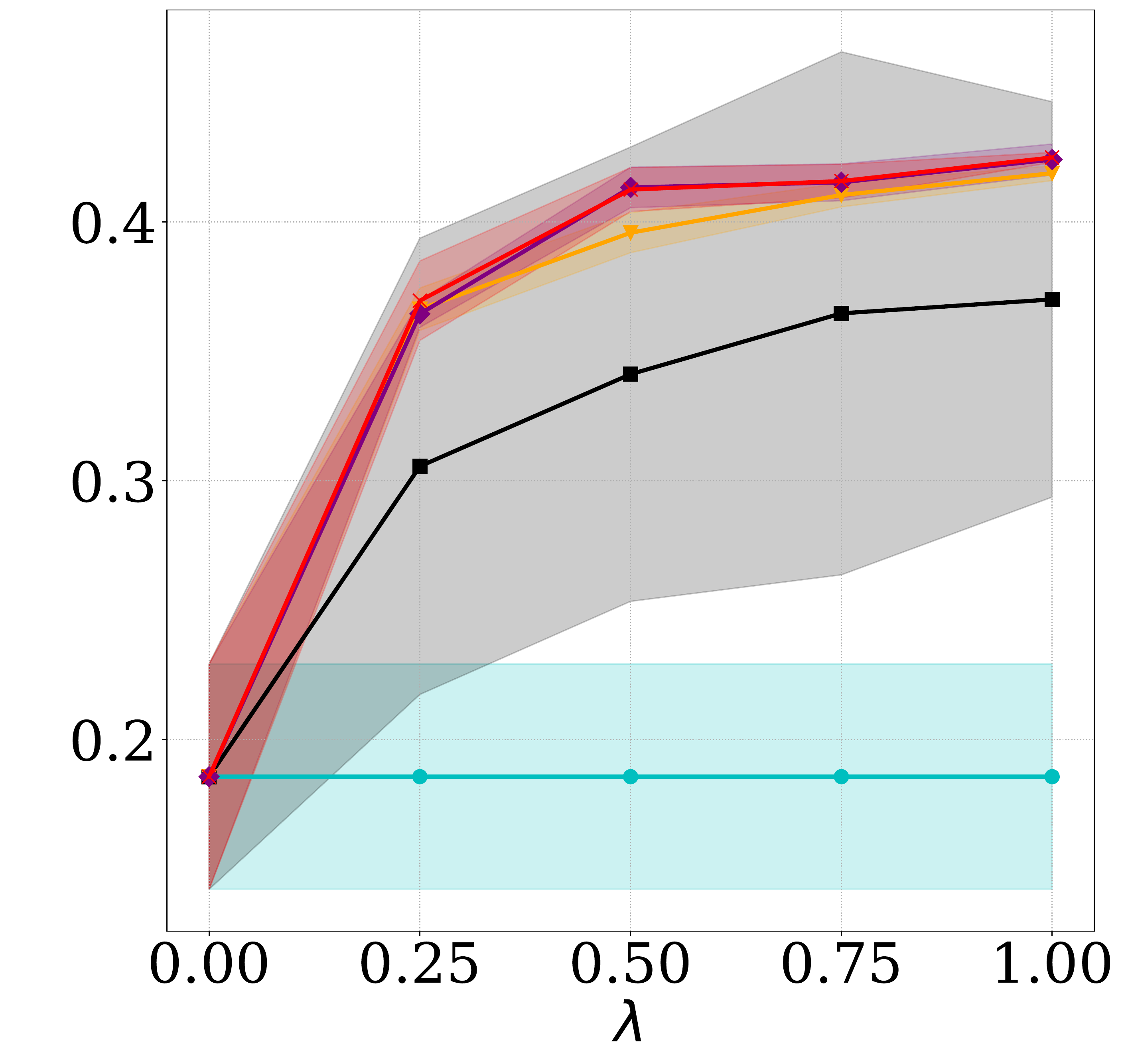}
            \hfill
            \includegraphics[width=0.49\textwidth]{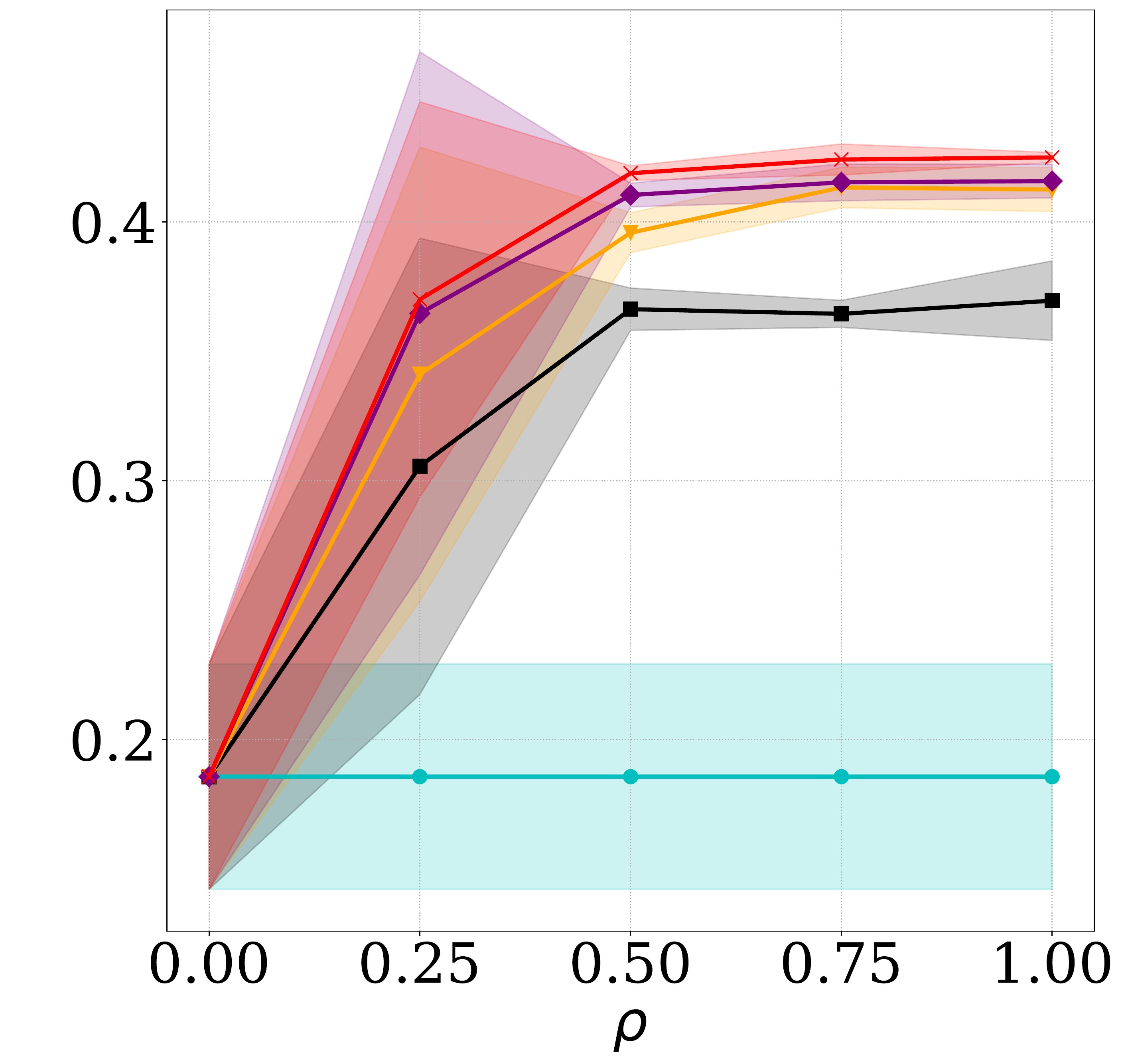}
            \caption{Coverage in Riverswim.}%
            \label{fig:coverage_riverswim}
        \end{subfigure}
        
        \begin{subfigure}[b]{\textwidth}   
            \centering 
            \includegraphics[width=0.49\textwidth]{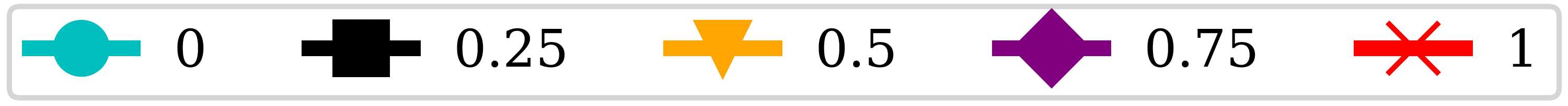}
        \end{subfigure}
        \caption{Coverage in LQG and Riverswim as function of $\lambda$ and $\rho$; average of 5 seeds, 95\% c.i..}
        \label{fig:coverage}
\end{figure*}
\begin{figure*}[htp]
        \centering
        \begin{subfigure}[b]{\textwidth}
            \centering
            \includegraphics[width=0.22\textwidth]{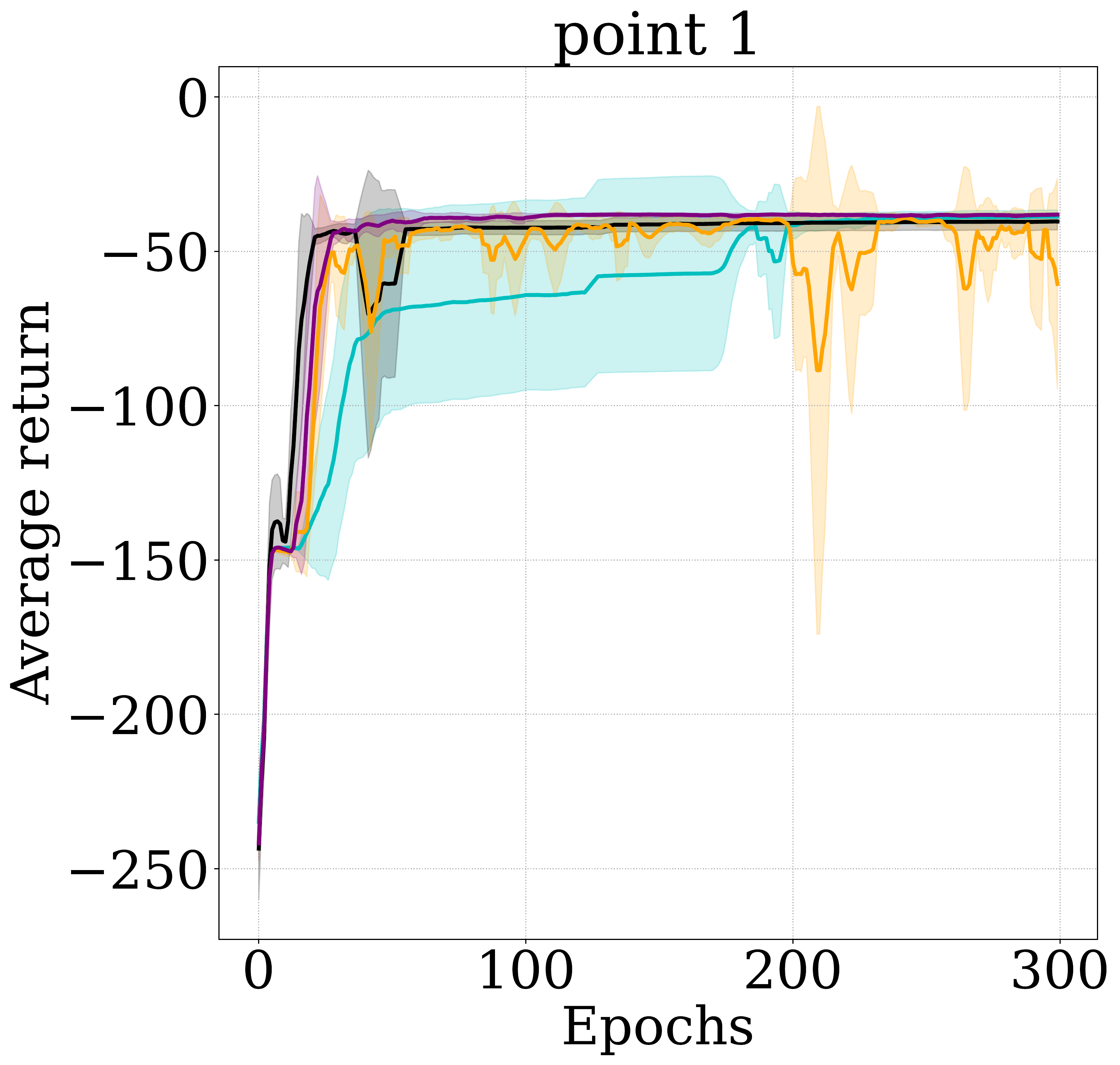}
            \hfill
            \includegraphics[width=0.22\textwidth]{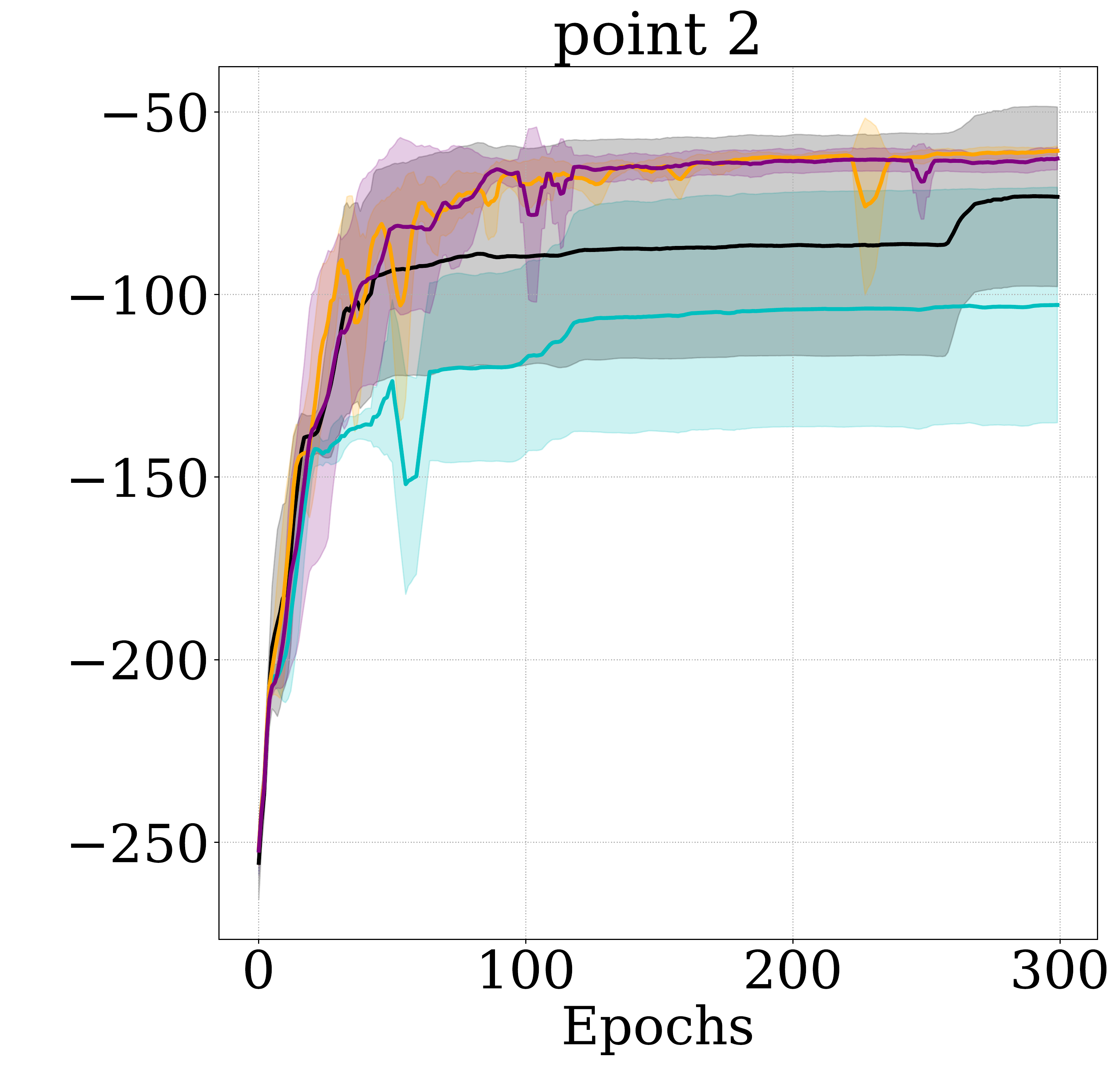}
            \hfill
            \includegraphics[width=0.22\textwidth]{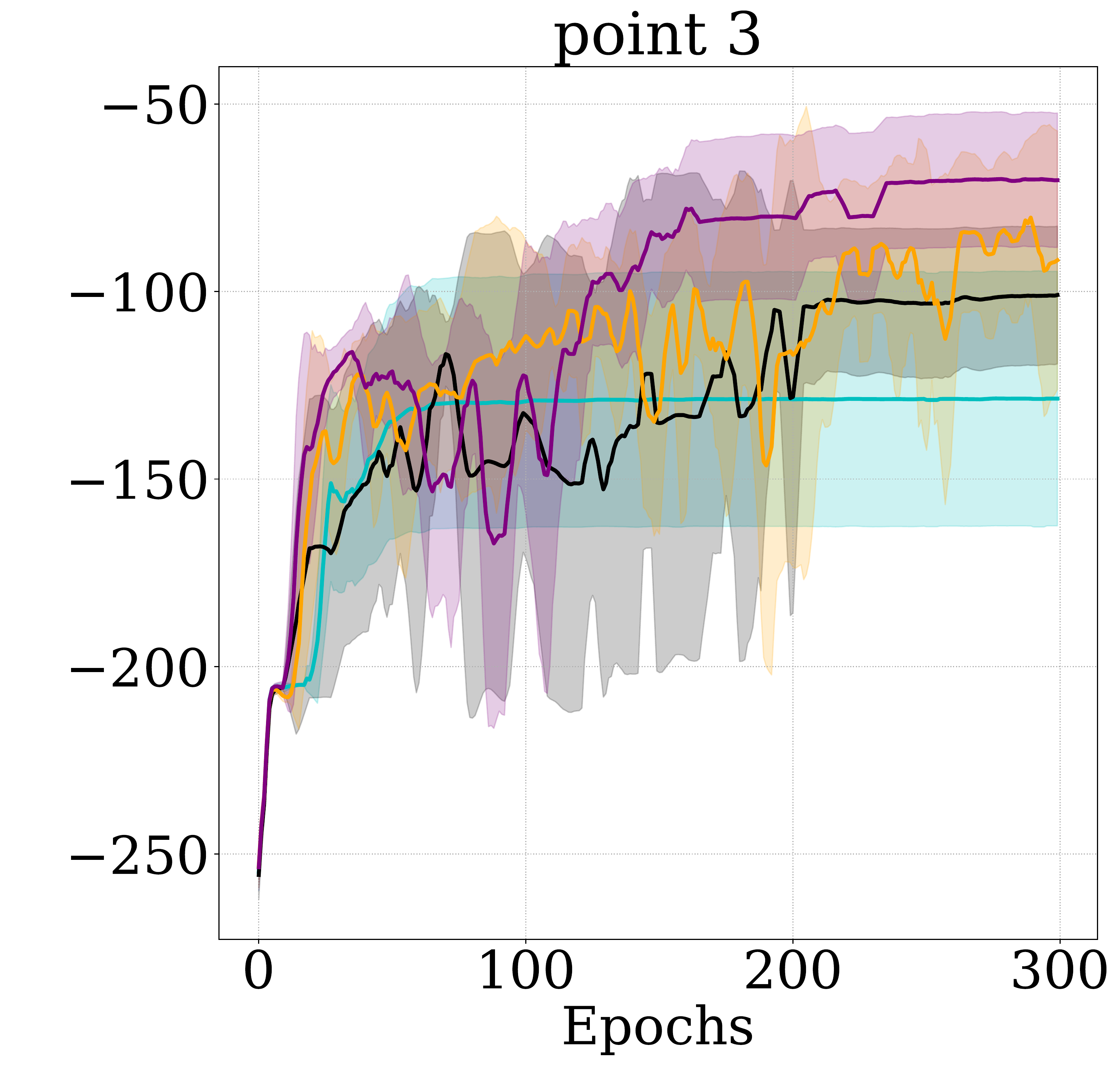}
            \hfill
            \includegraphics[width=0.22\textwidth]{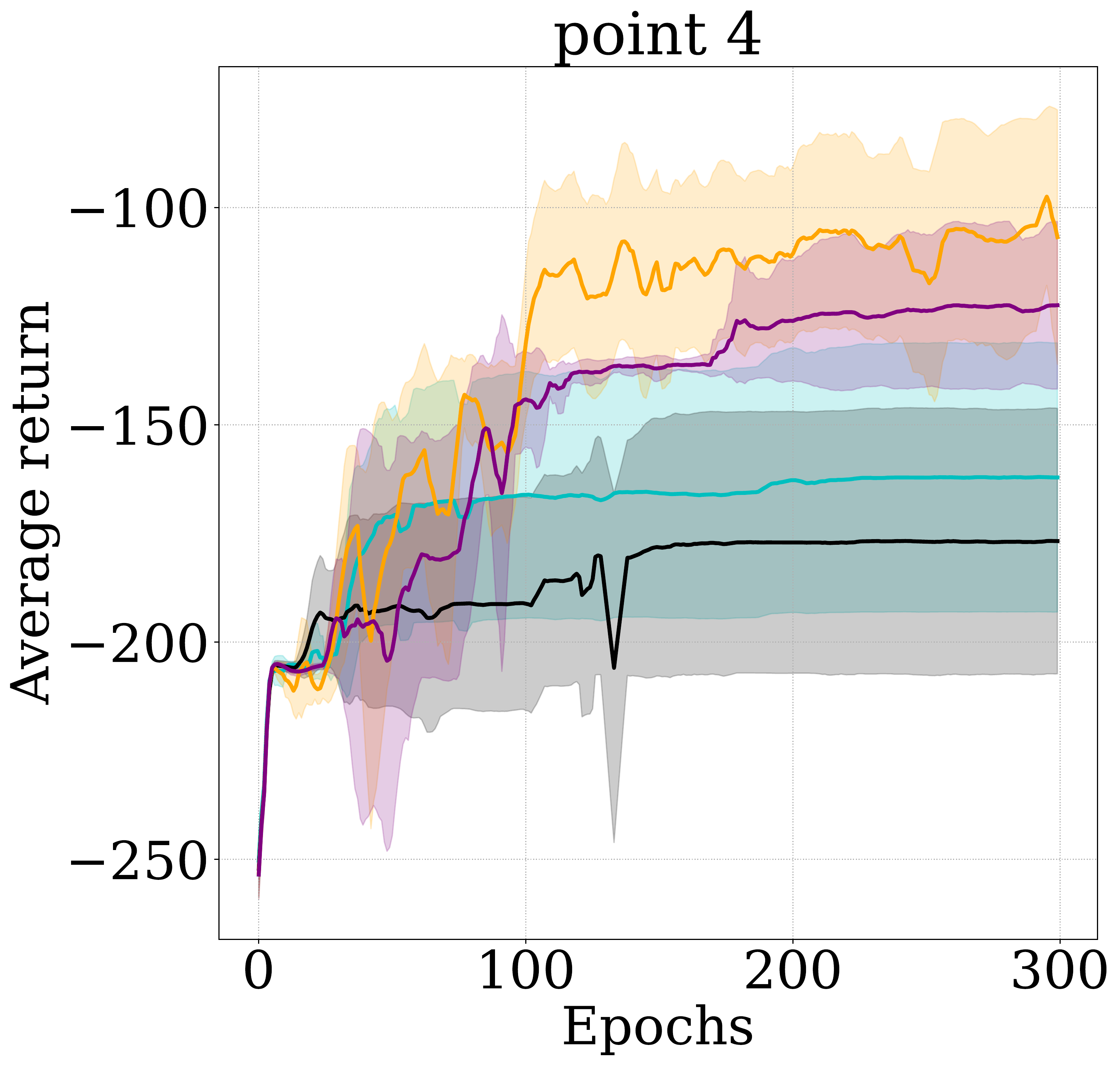}
            \caption{Average return in 4 2D navigation tasks.}
            \label{fig:point_distance_return}
        \end{subfigure}
        \\
        \begin{subfigure}[b]{\textwidth}   
            \centering
            \includegraphics[width=0.22\textwidth]{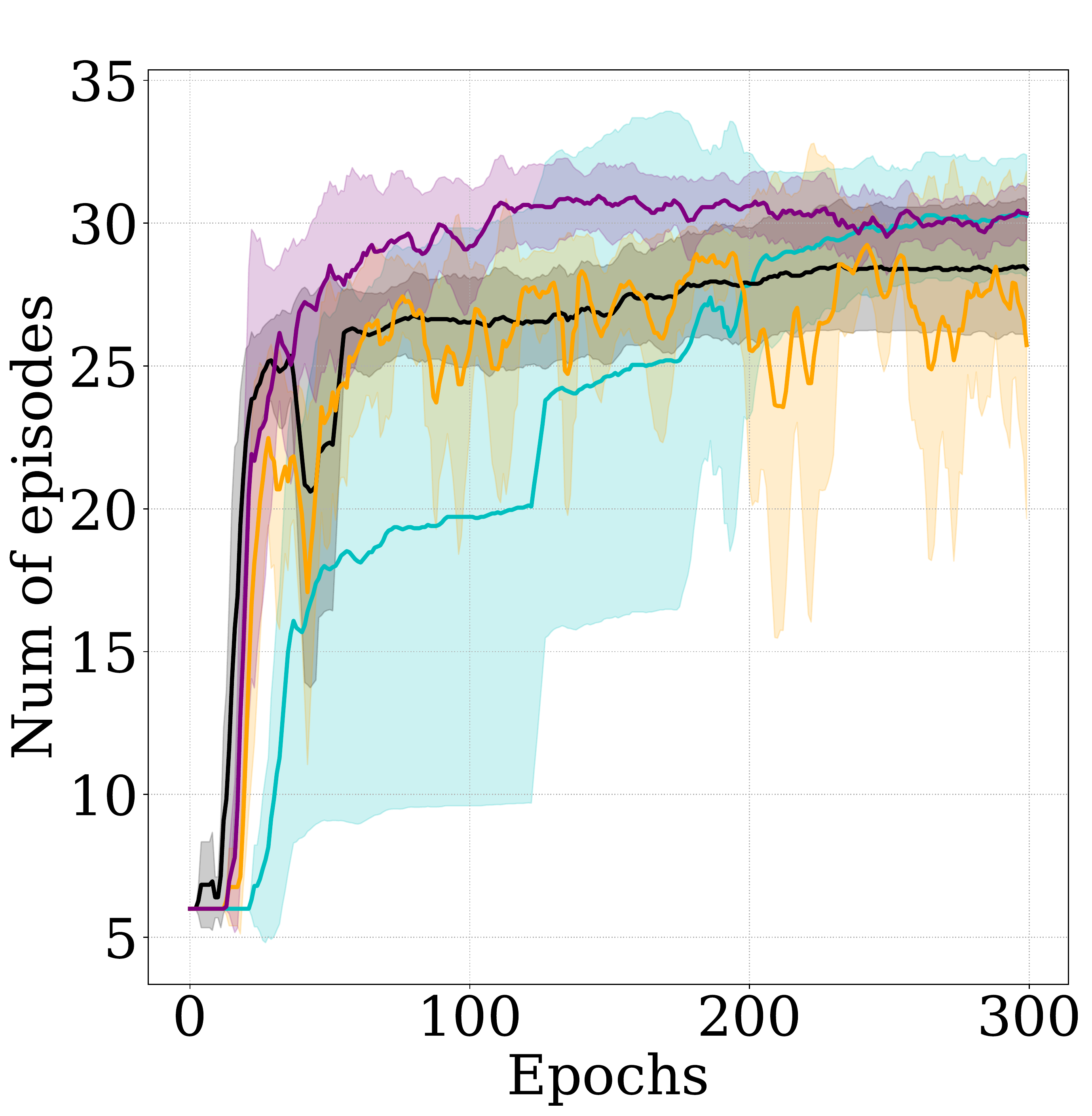}
            \hfill
            \includegraphics[width=0.22\textwidth]{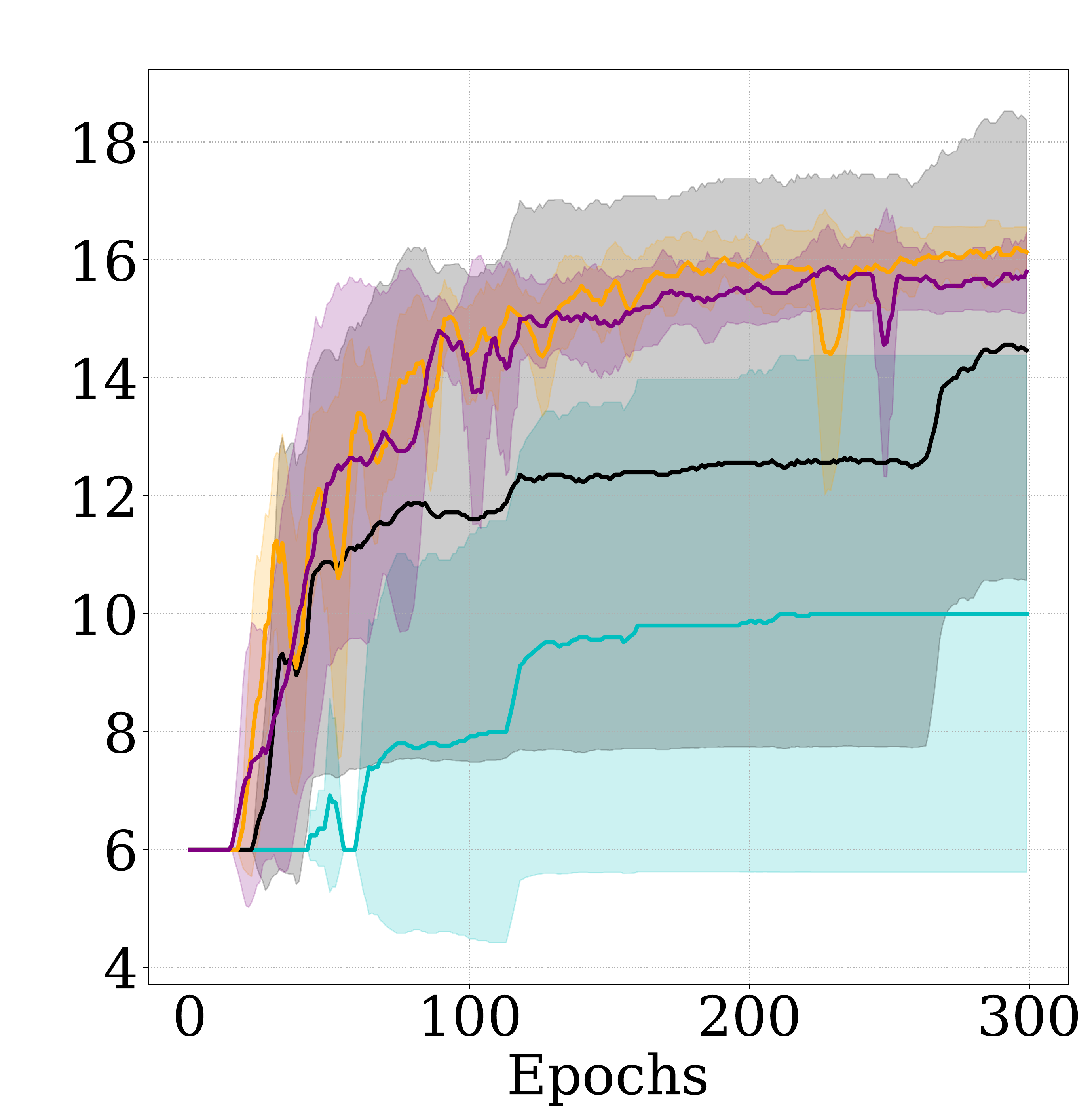}
            \hfill
            \includegraphics[width=0.22\textwidth]{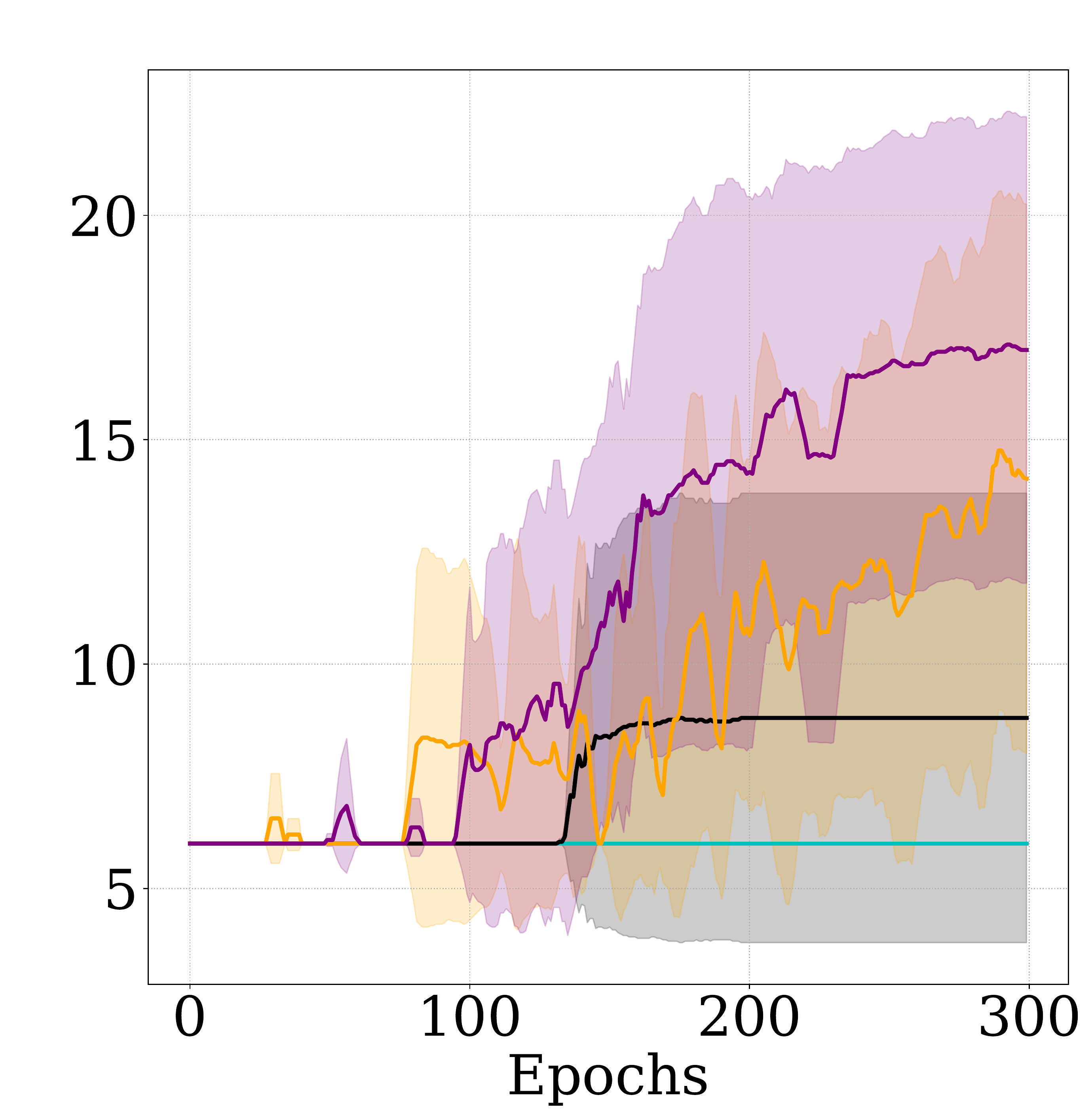}
            \hfill
            \includegraphics[width=0.22\textwidth]{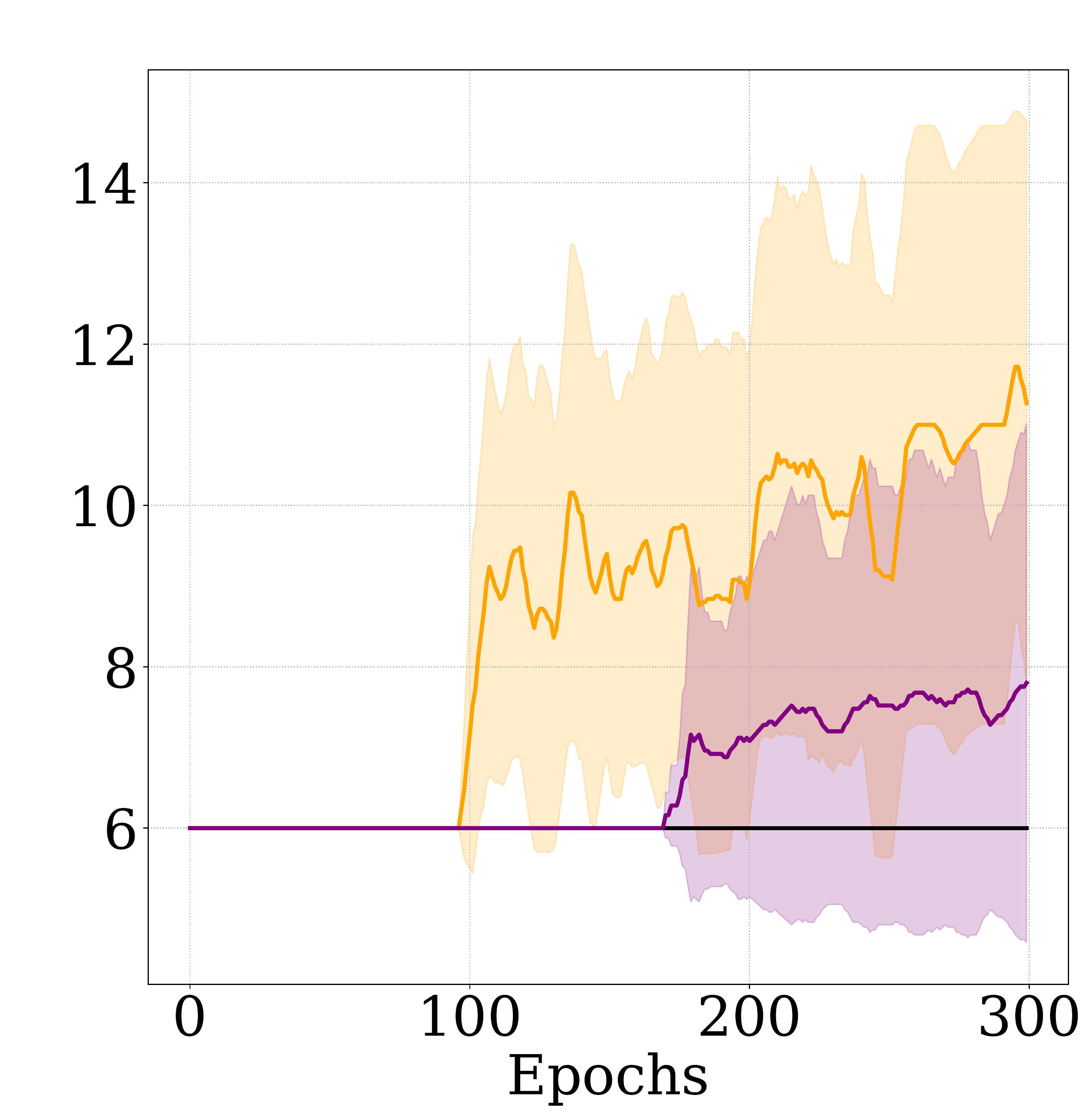}
            \caption{Number of episodes completed in 3000 steps in 4 2D navigation tasks.}
            \label{fig:point_distance_episodes}
        \end{subfigure}
        \\
        \begin{subfigure}[b]{\textwidth}   
            \centering
            \includegraphics[width=0.22\textwidth]{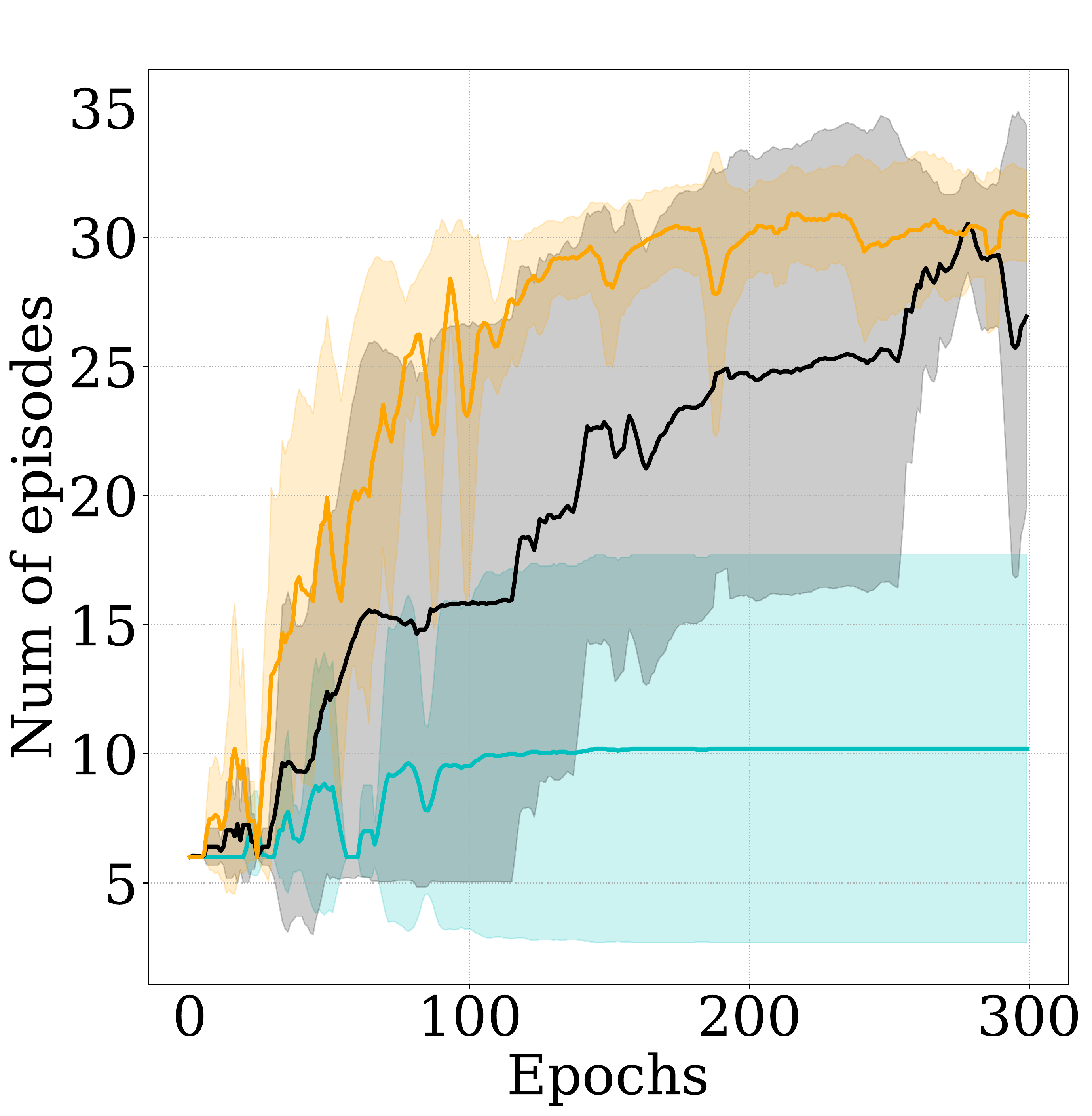}
            \hfill
            \includegraphics[width=0.22\textwidth]{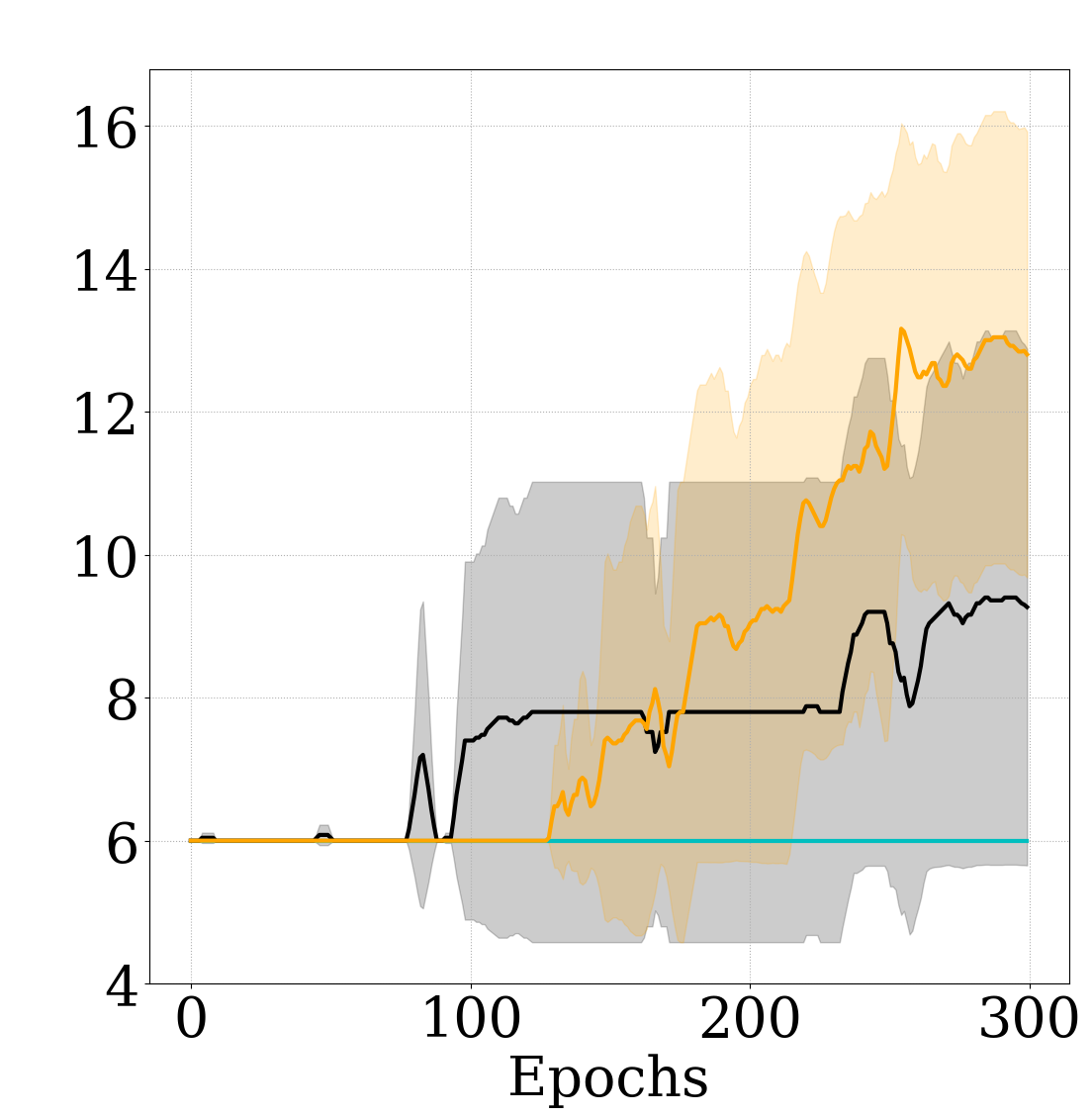}
            \hfill
            \includegraphics[width=0.22\textwidth]{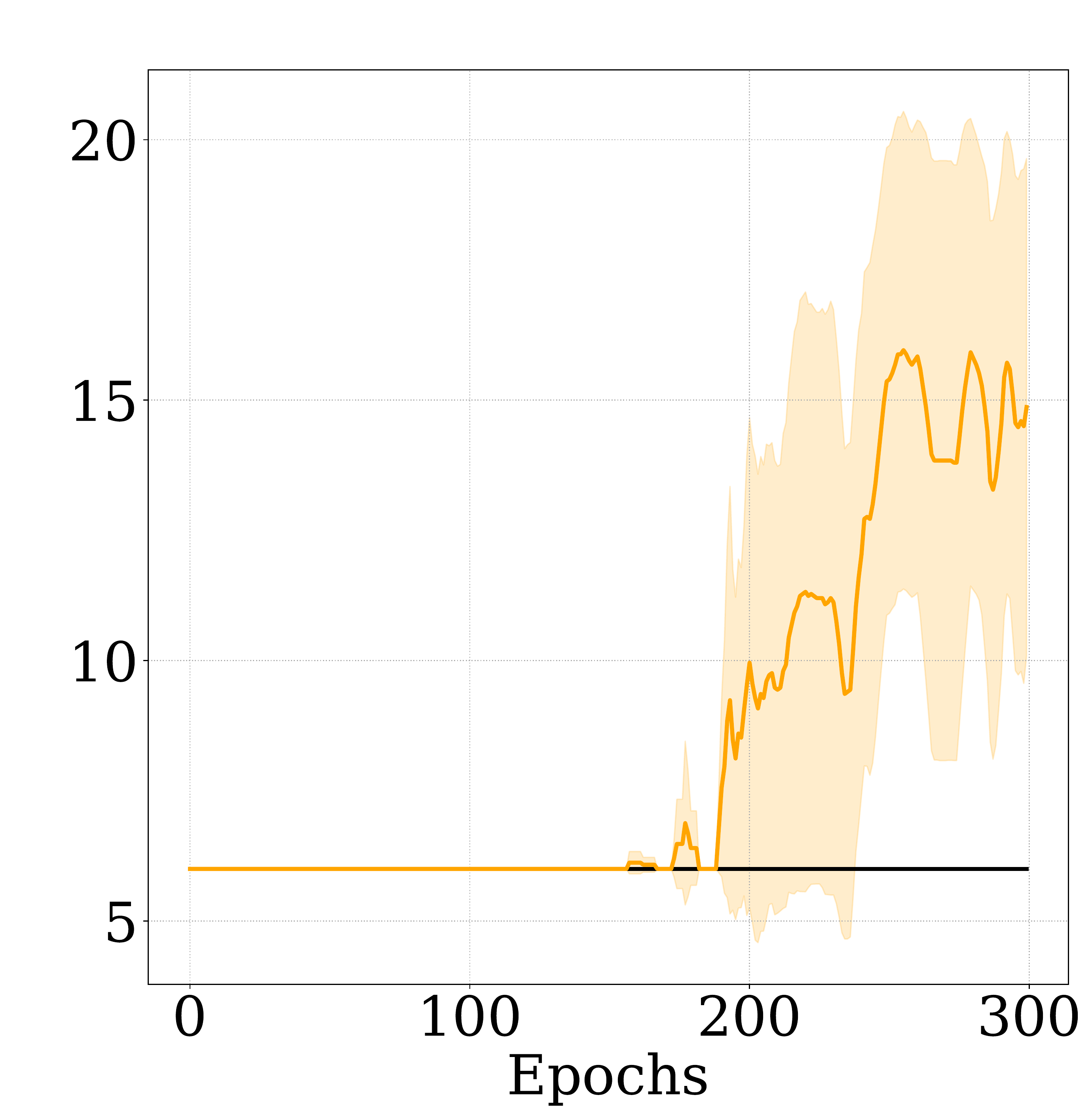}
            \hfill
            \includegraphics[width=0.22\textwidth]{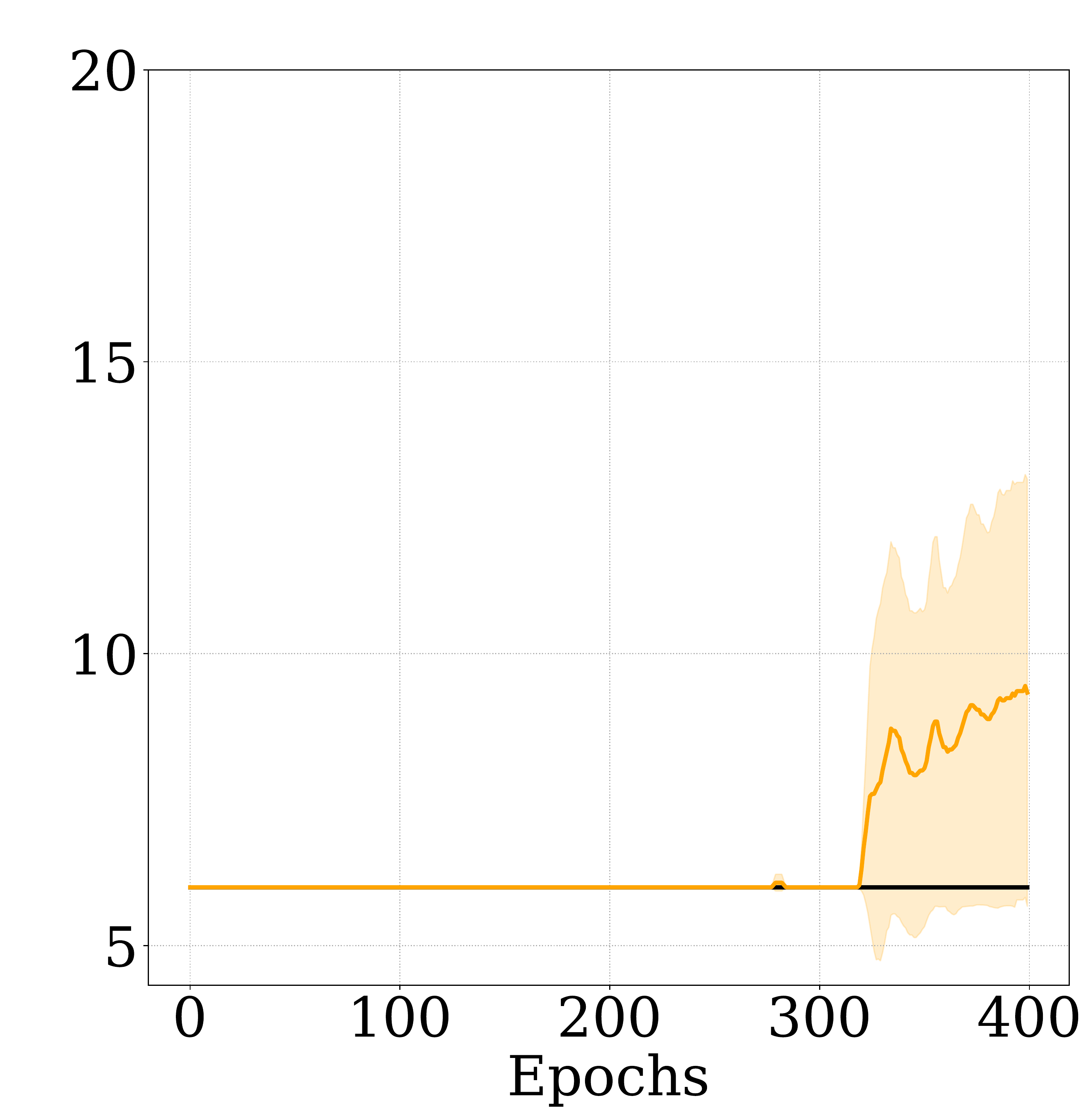}
            \caption{Number of episodes completed in 3000 steps in 4 2D navigation tasks with sparse reward function.}
            \label{fig:point_sparse_episodes}
        \end{subfigure}
        \begin{subfigure}[b]{\textwidth}   
            \centering 
            \includegraphics[width=0.49\textwidth]{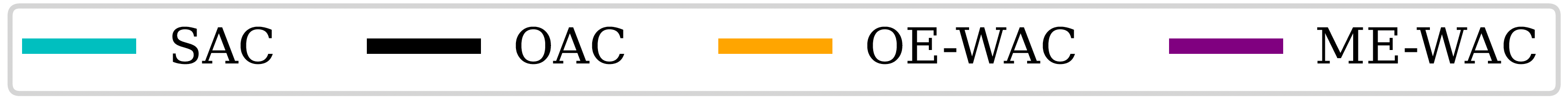}
        \end{subfigure}
        \caption{Experimental results in 4 2D navigation tasks starting from the easiest (left) to the hardest (right); average of 5 seeds, 95\% c.i..}
        \label{fig:point_distance}
\end{figure*}
Figure~\ref{fig:coverage} shows the results of these experiments. For each environment, we train WAC, varying the parameters $\lambda$ and $\rho$ of the regularized uncertainty loss in Equation~\eqref{eq:regularized_critic_loss}. For each value, we report the coverage averaged over all training epochs. 
Firstly, we observe that the coverage is monotonically increasing with both $\lambda$ and $\rho$. As expected, low values of $\rho$ cause higher variance, as fewer samples are employed to estimate the uncertainty regularization. This can be seen in all the curves of the leftmost plot, as well as in the third plot, where the black curve corresponding to $\rho=0.25$ suffers from a high variance. In Appendix B, 
we perform a similar study for OAC and we observe that the coverage is not so easily controllable. We attribute this to the heuristic nature of the uncertainty estimation of OAC. 

\paragraph{2D Navigation}
To assess whether a principled uncertainty estimation and propagation translate into lower sample complexity, we perform an empirical evaluation in a set of MujoCo~\citep{mujoco_todorov} tasks, where the amount of exploration needed to solve the task can be controlled. We start from the 2D navigation task used in~\citep{moro2022goaldirected}, where the agent has to reach a goal state in a 2D world, by avoiding obstacles. The reward is the negative Euclidean distance from the goal state. While this is a \emph{dense reward}, the obstacle presence generates local optima which the agent needs to overcome by exploring efficiently. We progressively make the task more challenging by 
adding additional walls and obstacles. We leave the full description of the environments in Appendix A.
 We name the tasks as \quotes{Point $x$} with $x \in\{1,2,3,4\}$, where a higher $x$ means a more difficult exploration challenge. We compare the performance of \algname, in both versions defined in Section~\ref{sec:wac}, with SAC and OAC. In each task, we track the cumulative return, as well as the number of episodes completed in a fixed number of steps (higher is better). We use the implementation of SAC and OAC used in~\citep{oac}, and extend the repository with our implementation, to guarantee comparable results. The same network architectures are used for all algorithms. 
For the common hyperparameters, we only tune SAC and use the same values for \algname and OAC, by additionally tuning the algorithm specific parameters ($\delta$ and $\beta$ for OAC and $\lambda$ and $\rho$ for \algname). Details on the hyperparameter tuning are in Appendix A.

In Figure~\ref{fig:point_distance_return}, we present the average return as a function of the training epochs, whereas in Figure~\ref{fig:point_distance_episodes} we present the number of episodes completed in 3000 steps of interaction. Starting from left to right, we increase the difficulty of the task. We can see that for the easiest task, all algorithms are able to find the optimal policy of quickly avoiding an obstacle in the middle to reach the goal state, even though SAC learns slower compared to the others. Being a simple exploration task, ME-WAC performs better and is more stable than OE-WAC. 
OAC is also able to quickly solve the task. While the difference in return is negligible, the number of completed episodes shows an advantage for ME-WAC, which completes more episodes faster. Finally, we underline that even though the task does not require particular exploration, WAC does not over-explore, but rather solves with a speed comparable with the other baselines. The clear advantages of WAC in terms of exploration can be seen starting from the second task, where the exploration requirements are increased. Both versions of WAC learn faster and with less variance compared with both SAC and OAC. The difference is even more apparent in the number of episodes completed, where SAC and WAC have disjoint confidence intervals. In the third task, SAC completely fails in learning to reach the goal, while OAC succeeds in some of the seeds only, showing a high variance. WAC, on the other hand, outperforms them both in terms of return and completed episodes. ME-WAC performs better, even though the task requires a good amount of exploration. Compared to ME-WAC, OE-WAC over-explores and it shows a slower learning curve. The last task is solved by the WAC agents only. SAC and OAC never reach the goal state. WAC outperforms them, in both versions, with statistical significance. We also see the need for larger exploration, apparent from the difference in performance between OE-WAC and ME-WAC. 

Finally, Figure~\ref{fig:point_sparse_episodes} presents the number of episodes completed in a \emph{sparse reward}
version of the same tasks. In this scenario, we do not show the return as it is proportional to the number of completed episodes. We only trained OE-WAC agents in these tasks, as they present a substantial exploration challenge. The advantage of WAC is extremely evident in these tasks.
SAC and OAC are only able to solve the simplest task. In a sparse reward setting, SAC and OAC will only explore randomly so they fully rely on the chance of reaching the goal state with random actions.
OAC explores more compared to SAC, but since the exploration does not depend on the state-action visitations, but only on the disagreement between the critics, sparse reward tasks are a great challenge. WAC, instead, will still explore,
even when facing sparse rewards since the uncertainty will gradually decline in visited regions, so the upper bounds will favor reaching unvisited ones. Indeed, OE-WAC outperforms both baselines in all the tasks with sparse rewards.

\section{Conclusions}
In this paper, we presented a novel AC algorithm to perform directed exploration. We presented \algname, which extends the recently proposed WQL to the continuous-actions case. Furthermore, we addressed a problem of uncertainty estimation that arises when using function approximation, related to the generalization of the uncertainty estimates. We proposed a simple, yet effective, regularization method, based on synthetic samples that allowed us to better generalize the uncertainty across the state-action space. Finally, we performed a thorough empirical evaluation to investigate the advantages of performing a principled uncertainty estimation and propagation in continuous-action domains. We observed that the uncertainty estimates of \algname can effectively steer exploration towards promising regions of the state-action space, even under sparse rewards, especially when comparing it with heuristic uncertainty estimation based on ensemble methods.
\bibliography{paper}

\onecolumn
\appendix
\section{Experimental Details}
\label{appx:a}
\subsection{Environments Description}

\textbf{RiverSwim}~~
We extend the classical Riverswim domain~\citep{strehl2008an} to a continuous setting. In this environment, the agent has to navigate a 1 dimensional state space, ranging from $0$ to $max\_state$, by applying a 1 dimensional action, representing the intended movement $a \in [-1 \quad 1]$. The initial state is a uniformly distributed in $[0 \quad 0.5]$. When an action is chosen the agent moves left or right on the state space. The distance of the movement is equal to the absolute value of the action (if the result is outside the state space a clip operations brings it back inside). The direction $d\in\{-1, 0, 1\}$ of the movement is stochastic, according to the following probabilities:

\begin{equation*}
\label{eq:riverswim_probs}
\begin{aligned}
P(d=-1 | a) &= \begin{cases}
   1 - 0.9 \cdot (a + 1)  &\text{if } a \leq 0 \\
   0.1 &\text{if } a > 0
\end{cases}\\
P(d=0 | a) &= \begin{cases}
   0.9 \cdot (a + 1)  &\text{if } a \leq 0 \\
   0.9 - 0.3 a &\text{if } a > 0
\end{cases}\\
P(d=1 | a) &= \begin{cases}
   0  &\text{if } a \leq 0 \\
   0.3a  &\text{if } a > 0
\end{cases}\\
\end{aligned}
\end{equation*}
Given the current state $s_t$, the action $a_t$ and the direction $d_t$ sampled according the previous probabilities, the next state is $s_{t+1} = \text{clip}(s_t + d_t |a_t|)$, where $\text{clip}$ clips the state in the range $[0 \quad max\_state]$
The reward depends on the starting state $s$ and the action sign:

\begin{equation*}
r = \begin{cases}
    5 \cdot 10^{-4} &\text{if } s \leq 1 \\
    1 &\text{if }  s \geq (max\_state - 1) \text{ and } a > 0 \\
    0 &\text{otherwise}  \\
\end{cases}
\end{equation*}

The optimal policy is to always perform $a=1$ which gives the agent the best chance of moving toward high reward states.

In our experiments: 
\begin{itemize}
  \item $max\_state = 25$
\end{itemize}

\textbf{LQG}~~ We test our agents also on an instance of a Linear Quadratic Gaussian control. Given a state $x$, an action $a$, and $v \sim \mathcal N(0,0.5)$ the next state and the cost $c\ (=-r)$ are defined as: 

\begin{equation*}
\begin{aligned}
x' &= Ax + Ba + v \\
c &= Qx^2 + Ra^2 
\end{aligned}
\end{equation*}

In our experiments, we use  $A = 1, B = 1, Q = 0.9, R = 0.9$. The agents starts from the borders of the state space and, with this configuration, the goal is to reach the origin of the state space while balancing the actions.

\textbf{Point}~~
This environment models a sphere moving inside a two-dimensional maze. The goal of the agent is to get close enough to a goal state on the right side of the maze, while avoiding obstacles. Once the agent gets close enough to the goal (Euclidean distance $<2$) the episode ends. The state space includes the agent position in the $2$-dimensional space, as well as the velocities. The action space is also $2$-dimensional, controlling the actuators in both directions. We use 2 reward functions for this task. In the dense reward version of the environment, the reward is the negative Euclidean distance of the sphere from the center of the goal. This represents a dense reward signal, which  makes optimization easier but also introduces local maximums due to the presence of the obstacles. In the sparse version the reward is always $-1$, so the optimal policy is to reach the goal as quickly as possible so that the episodes ends. We devise 4 environment configurations, with different levels of difficulty.

\begin{figure}
        \centering
        \begin{subfigure}[b]{0.49\textwidth}
            \centering
            \includegraphics[width=0.8\textwidth]{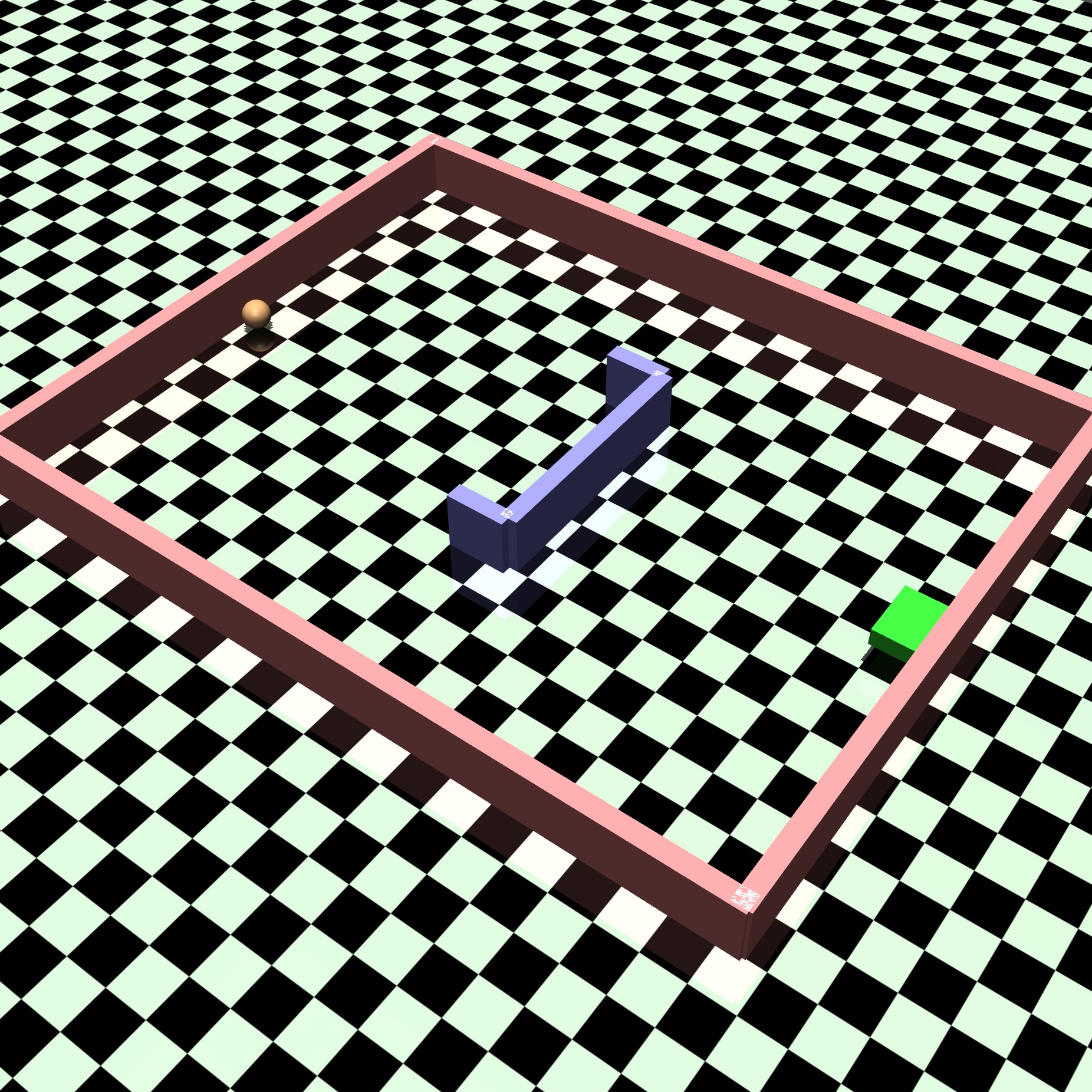}
            \caption{Point 1}
            \label{fig:point_1}
        \end{subfigure}
        \hfill
        \begin{subfigure}[b]{0.49\textwidth}   
            \centering 
           \includegraphics[width=0.8\textwidth]{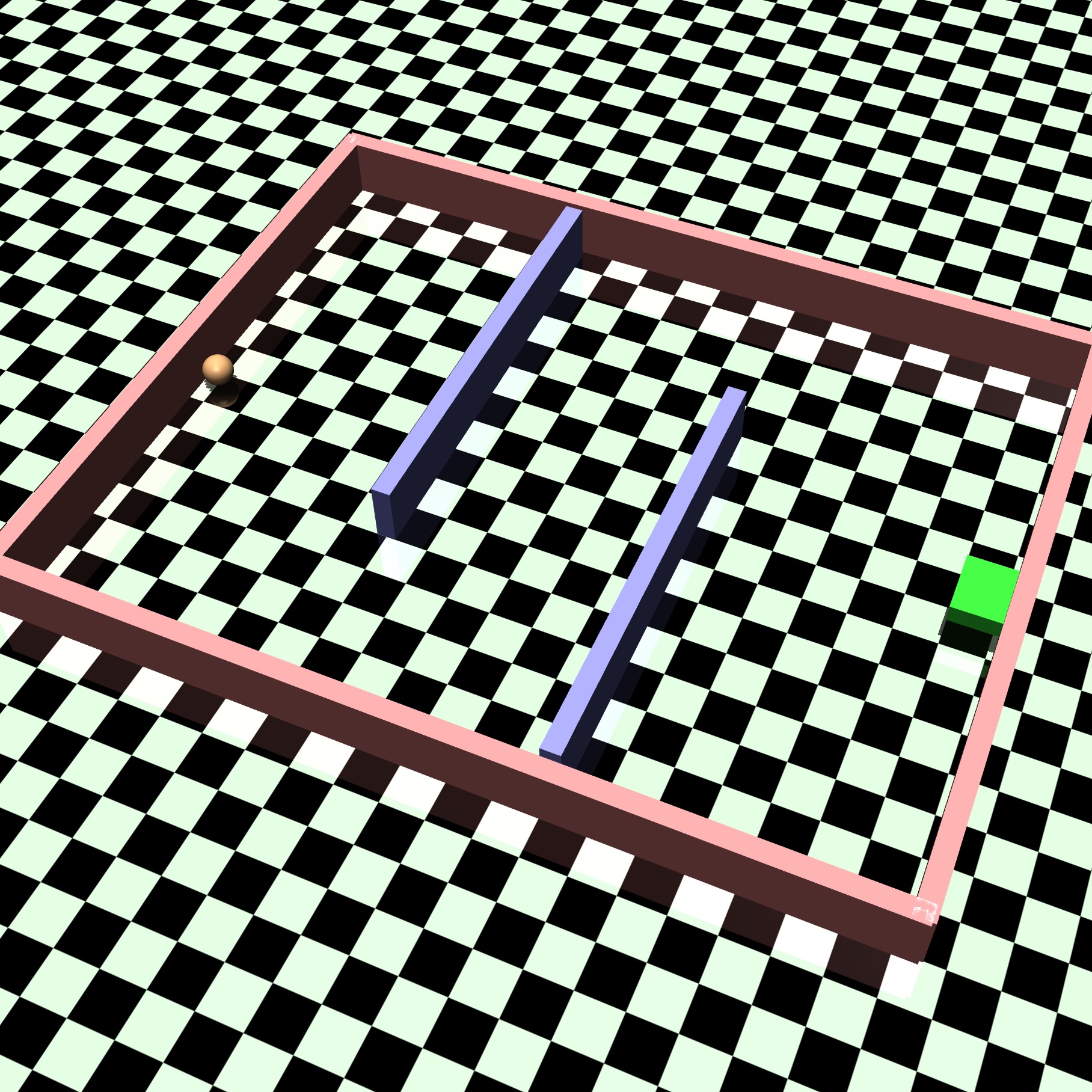}
            \caption{Point 2}%
            \label{fig:point_2}
        \end{subfigure}
                \begin{subfigure}[b]{0.49\textwidth}
            \centering
            \includegraphics[width=0.8\textwidth]{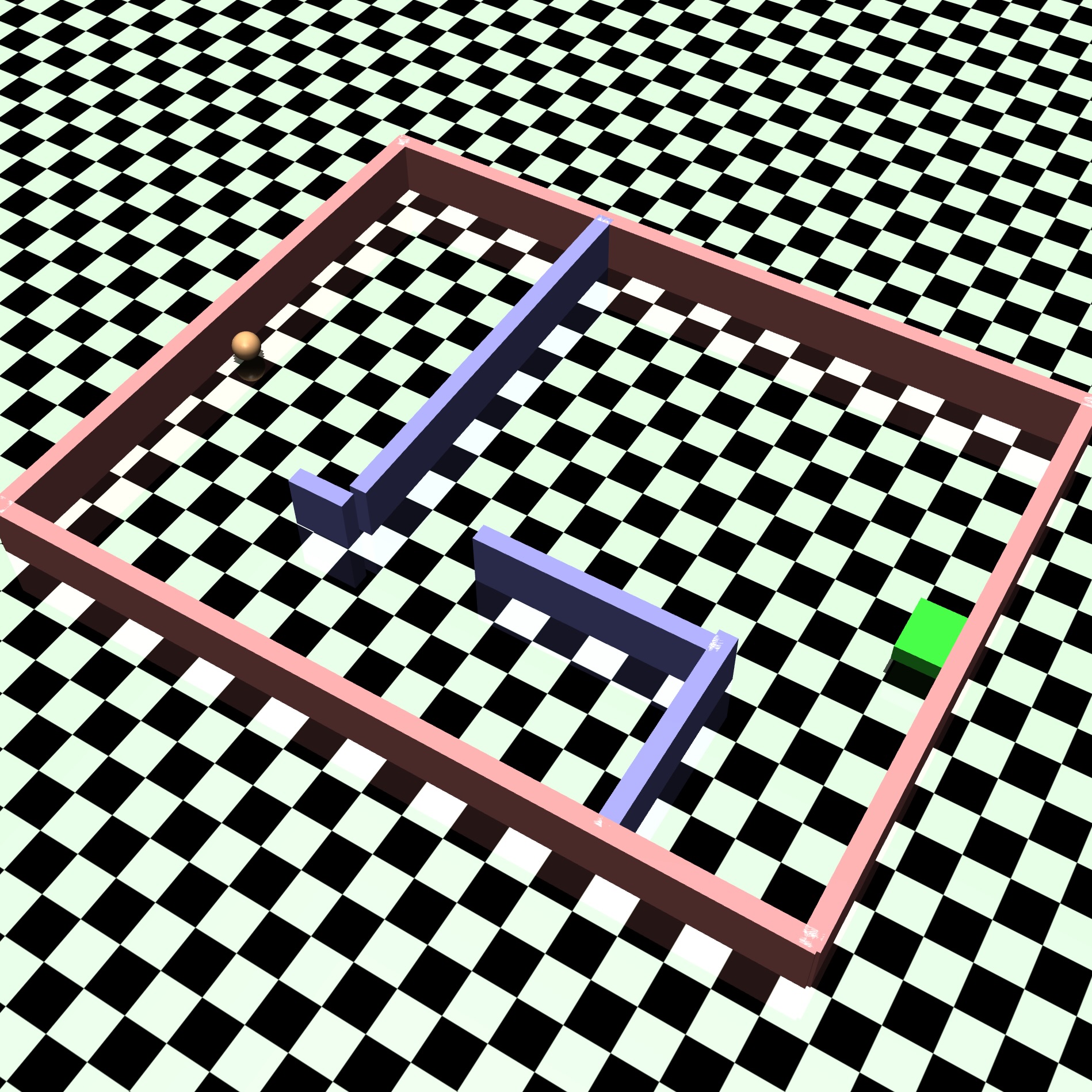}
            \caption{Point 3}
            \label{fig:point_3}
        \end{subfigure}
        \hfill
        \begin{subfigure}[b]{0.49\textwidth}   
            \centering 
           \includegraphics[width=0.8\textwidth]{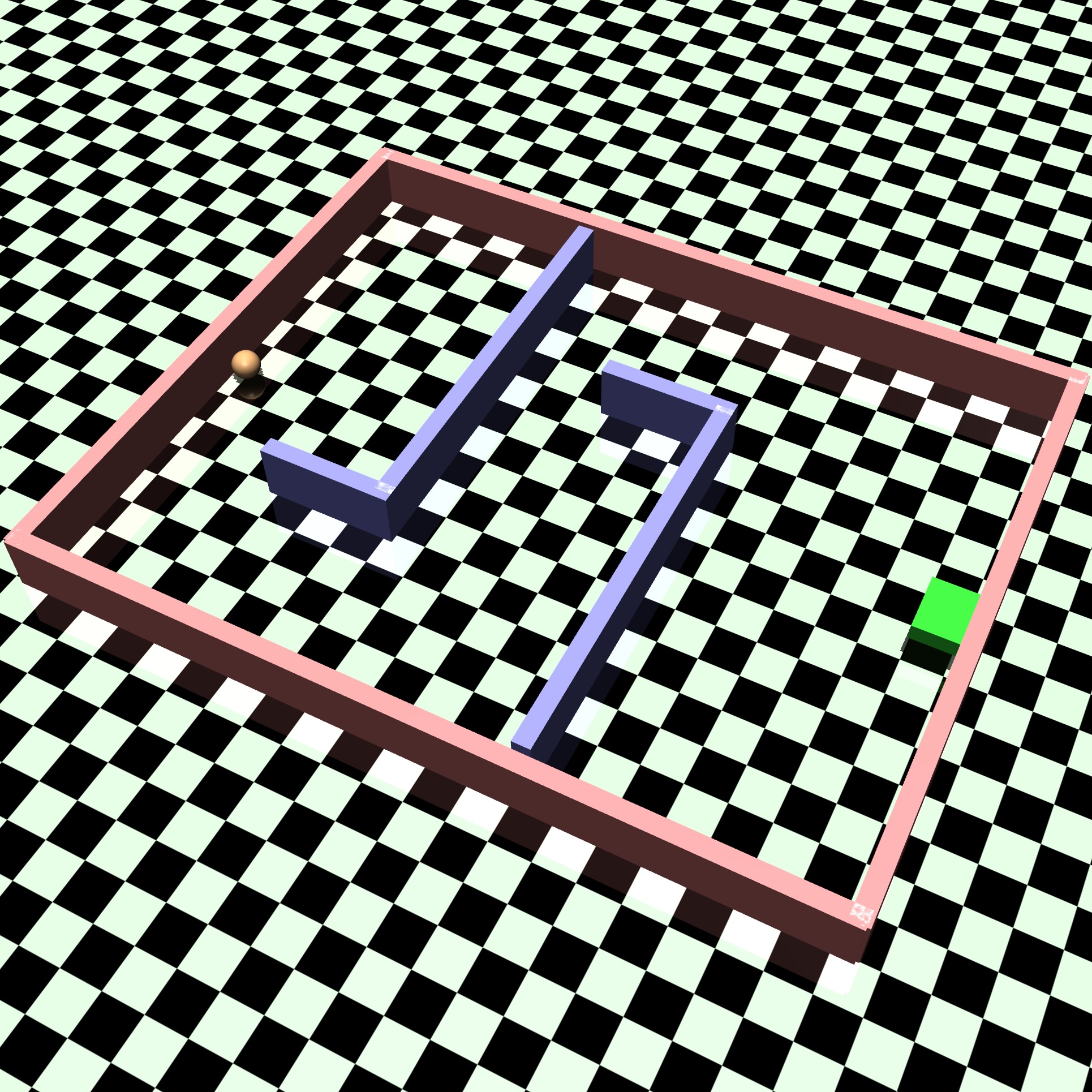}
            \caption{Point 4}%
            \label{fig:point_4}
        \end{subfigure}
        \label{fig:point_envs_description}
\end{figure}

The first environment (Figure~\ref{fig:point_1}) has a U-shaped wall in the middle. The agent has to overcome it by either running into it and then moving back to outflank it, so it can escape the local maximum, or preferably, it has to go around it without touching it. The U-shaped of the wall makes the task more difficult since the agent is unlikely to be able to escape the local maximum once it has hit the wall by simply playing random actions. 

The second environment (Figure~\ref{fig:point_2}) has $2$ simpler obstacles to overcome compared to first environment, yet the agent now has to learn how to overcome two obstacles which overall makes the second task more difficult than the first one.

To overcome the obstacle of the third environment (Figure~\ref{fig:point_3}) the agent has to perform a rather complete exploration of the space since the gateway to the second room is quite narrow and it is easy to be stuck in a local maximum at the border due to the reward function based on the Euclidean Distance.

The forth and last environment shown in Figure~\ref{fig:point_4}, puts together the challenges of the previous two, it has 2 obstacles to overcome and it also requires a rather complete exploration of the space to advance to the goal.

\subsection{Tuning procedures}
In this appendix, we present the hyperparameter tuning employed, as well as the final values used in our experiments. For all the approximators employed, including critics and actors, we use 2 layer MLPs.
For what concerns LQG all algorithms could solve it easily independently from the hyper-parameters chosen. The same can be said for RiverSwim, except for SAC which could not solve for any set of hyper-parameters contained in the grid search we performed. 

In point environment we performed a grid search on all environments with dense rewards starting with SAC (3 runs with 3 different seeds for each node of the grid). We choose the set of hyper-parameters that could solve the most runs for the two most difficult environments in which at least one run could learn a policy that reached the goal. The best recovered values are reported in Table~\ref{table:SACtable}.

\begin{table}[ht]
\caption{SAC parameters} 
\centering 
\begin{tabular}{l c} 
\hline\hline 
\textbf{Parameter} & \textbf{best value} \\ [0.5ex] 
\hline 
networks' number of layers & $2$ \\ 
layers' size & $256$ \\
replay buffer size & $10^6$ \\
number of train steps per train loop & $1000$ \\
number of exploration steps per train loop & $1000$ \\ 
batch size & $256$ \\
learning rates & $10^{-3}$ \\
\hline 
\end{tabular}
\label{table:SACtable} 
\end{table}

Afterwards, we performed a grid search on OAC and WAC, where we fixed all the hyper-parameters they share with SAC to the best values we found on SAC hyper-parameter tuning and we tuned only on their additional hyper-parameters. Once again, we choose the hyper-parameters sets that allow each algorithm to perform best the 2 most difficult environment it could solve. The final values are reported in Table~\ref{table:OACtable} and Table~\ref{table:WACtable}.


\begin{table}[!htb]
\begin{minipage}{.5\linewidth}
    \centering

    \caption{OAC parameters}
    \label{table:OACtable} 

    \medskip

\begin{tabular}{l c} 
\hline\hline 
\textbf{Parameter} & \textbf{best value} \\ [0.5ex] 
\hline 
$\delta$ & $18$ \\ 
$\beta_{UB}$ & $6.5$ \\
\hline 
\end{tabular}
\end{minipage}\hfill
\begin{minipage}{.5\linewidth}
    \centering

    \caption{WAC parameters}
    \label{table:WACtable} 

    \medskip

\begin{tabular}{l c} 
\hline\hline 
\textbf{Parameter} & \textbf{best value} \\ [0.5ex] 
\hline 
$\delta$ & $0.95$ \\ 
$\lambda$ & $0.6$ \\
$\rho$ & $0.6$ \\
\hline 
\end{tabular}
\end{minipage}
\end{table}



\section{Additional Experiments}
\label{appx:b}

\subsection{Standard MujoCo experiments}
The environments we employed for evaluation are characterized by relevant exploration challenges. This choice, in our view, is motivated by the fact that we propose an approach to effectively address exploration in continuous state-action environments. We decided not to include evaluations on standard MuJoCo tasks in the main paper, as they do not pose great exploration challenges. For example, we empirically investigated in our paper some flaws with the exploration method employed by OAC and the original OAC paper only performs evaluation in standard MuJoCo tasks, that is why we move away from those tasks. Nonetheless, for completion, we present here some additional evaluations in the environments of HalfCheetah-v2, Hopper-v2 and Walker2d-v2.

\begin{figure}
        \centering
        \begin{subfigure}[b]{0.9\textwidth}
            \centering
            \includegraphics[width=0.5\textwidth]{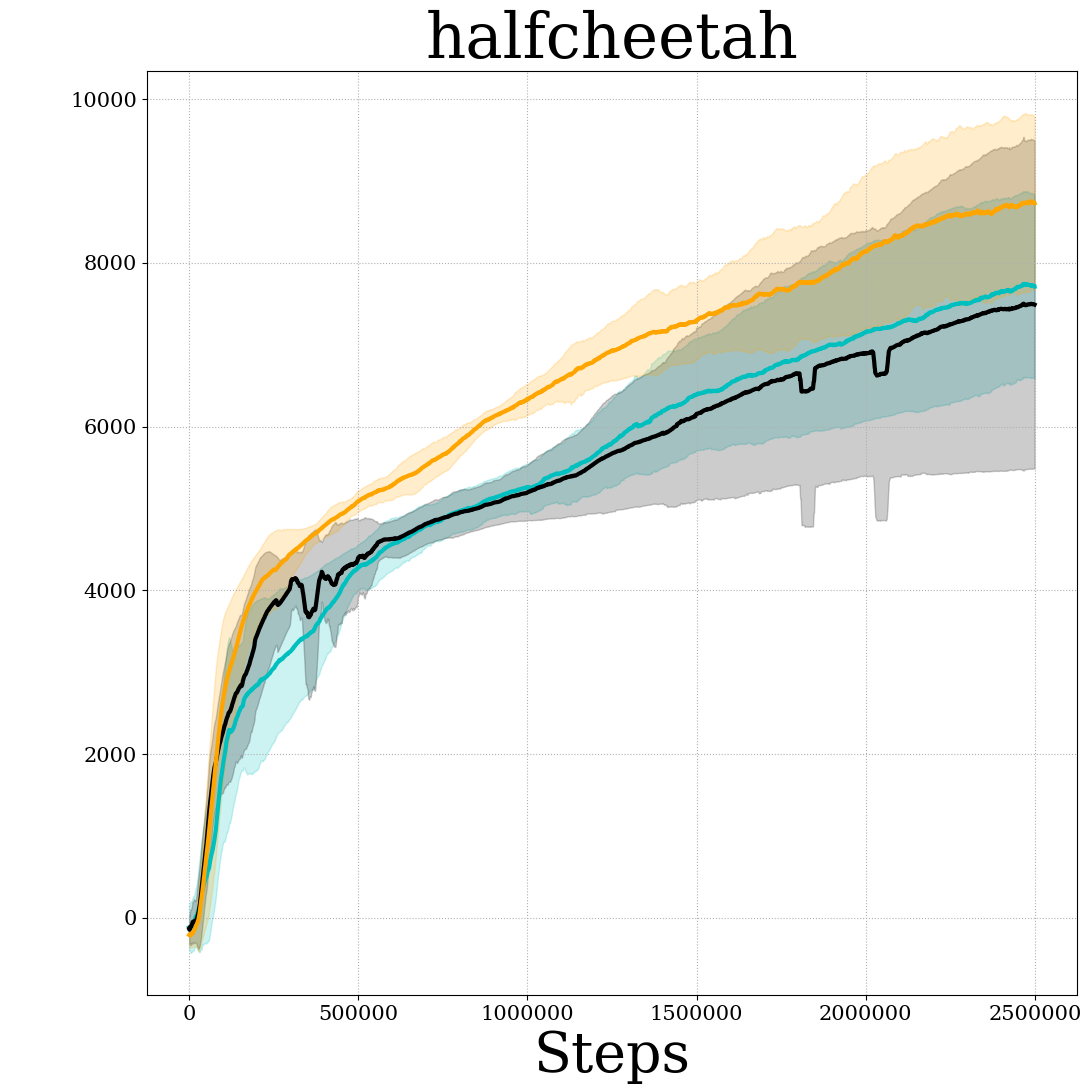}
            \label{fig:mujoco_cheetah}
        \end{subfigure}
        \hfill
        \begin{subfigure}[b]{0.9\textwidth}   
            \centering 
           \includegraphics[width=0.5\textwidth]{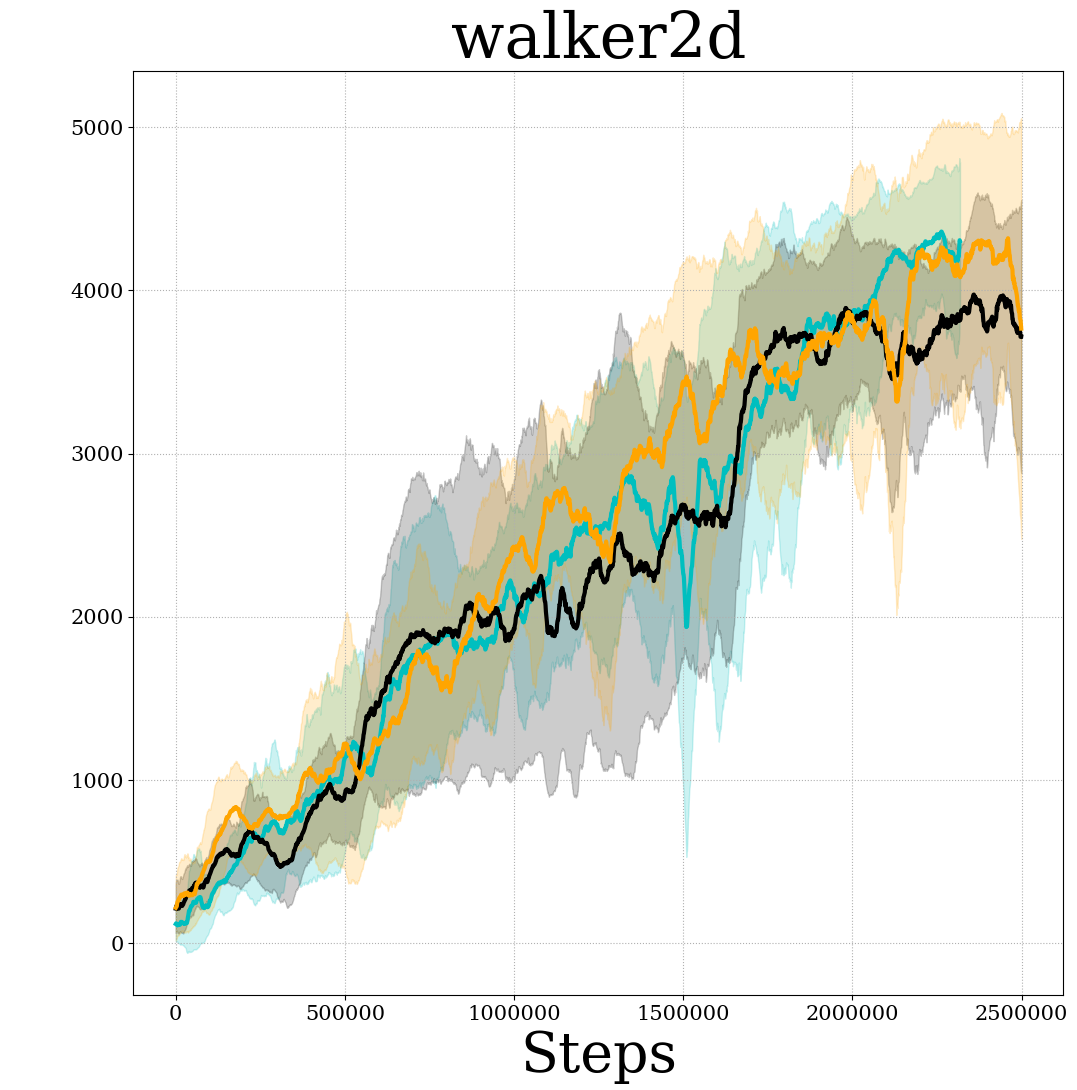}
            \label{fig:mujoco_walker}
        \end{subfigure}
                \begin{subfigure}[b]{0.9\textwidth}
            \centering
            \includegraphics[width=0.5\textwidth]{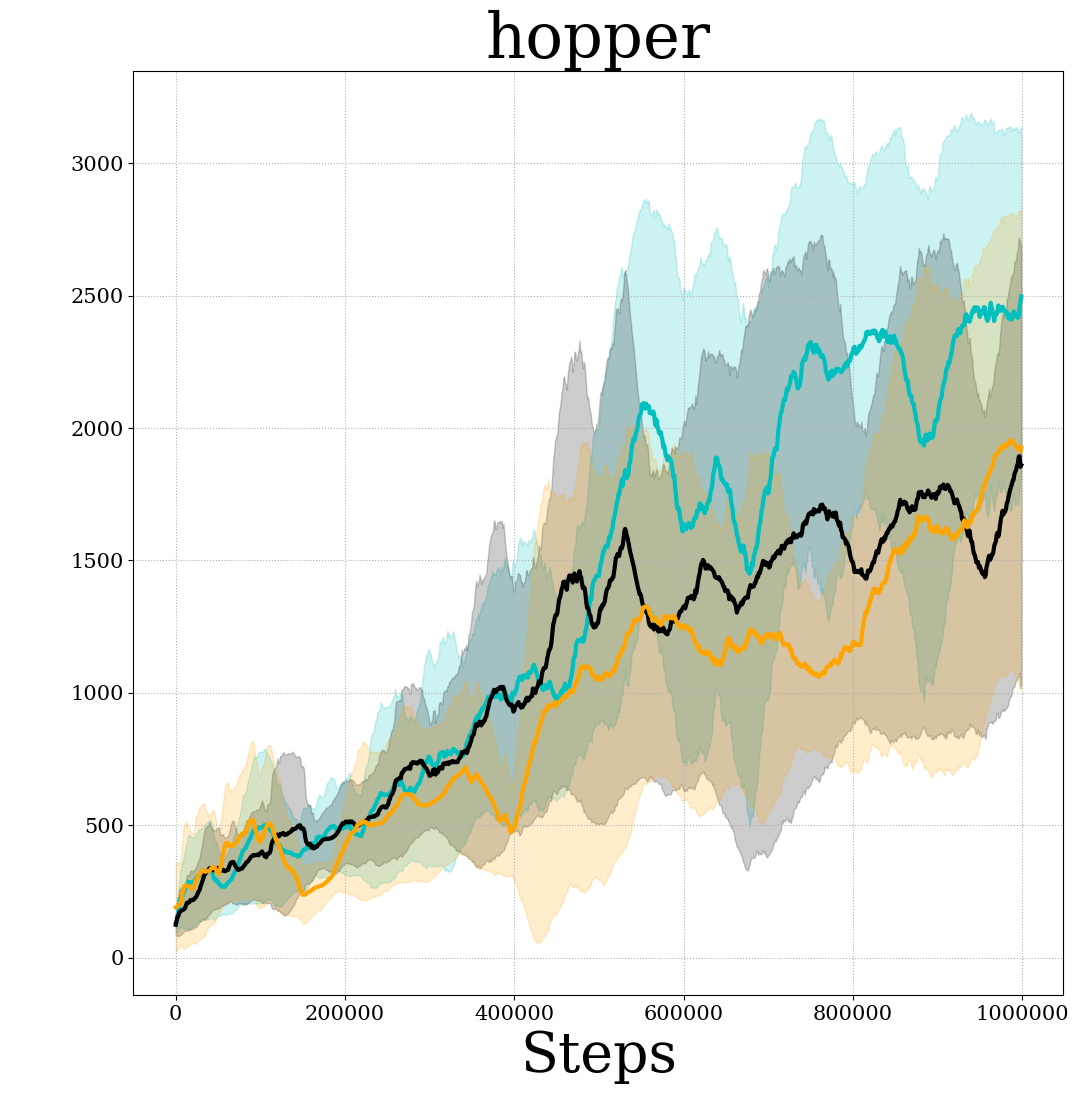}
            \label{fig:mujoco_hopper}
        \end{subfigure}
        \hfill
        \begin{subfigure}[b]{0.9\textwidth}   
            \centering 
           \includegraphics[width=0.5\textwidth]{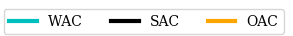}
        \end{subfigure}
        \caption{Empirical evaluation in 3 standard MujoCo tasks. Average of 4 seeds, 95 \% c.i..}
        \label{fig:mujoco_results}
\end{figure}

The results are shown in Figure~\ref{fig:mujoco_results}. We do not perform tuning on the WAC exploration parameters, whereas OAC and SAC are run with the (tuned) parameters reported in the OAC paper. We are outperformed by OAC in HalfCheetah-v2, outperform both SAC and OAC in Hopper-v2 and perform the same in Walker2d-v2. This supports our claim that standard MujoCo tasks are not suitable benchmarks for exploration.

\subsection{Tuning on $\delta$}

\begin{figure}
        \centering
        \begin{subfigure}[b]{0.49\textwidth}
            \centering
            \includegraphics[width=1\textwidth]{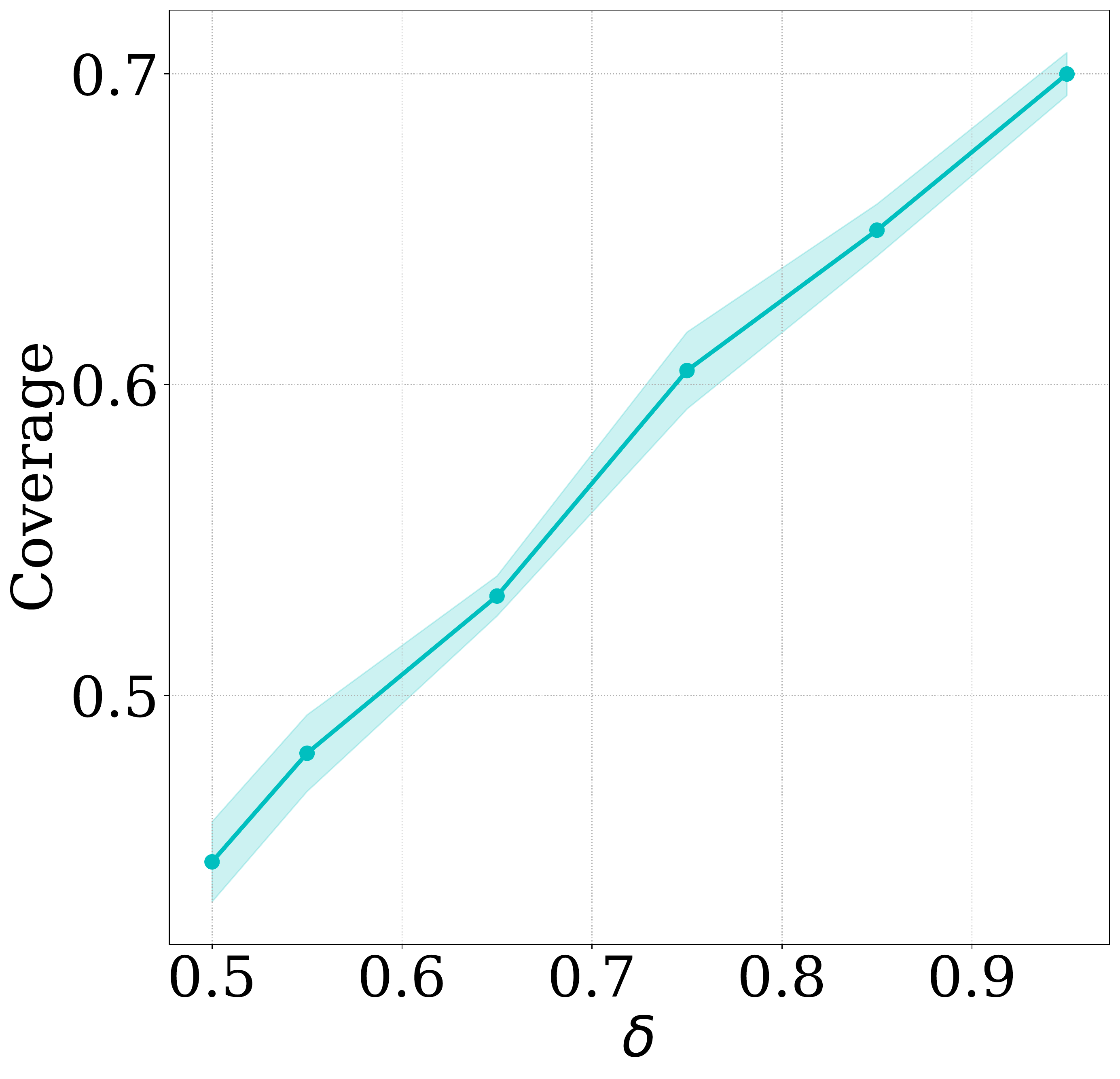}
            \caption{Coverage in LQG.}
            \label{fig:covergae_lqg_delta}
        \end{subfigure}
        \hfill
        \begin{subfigure}[b]{0.49\textwidth}   
            \centering 
            \includegraphics[width=1\textwidth]{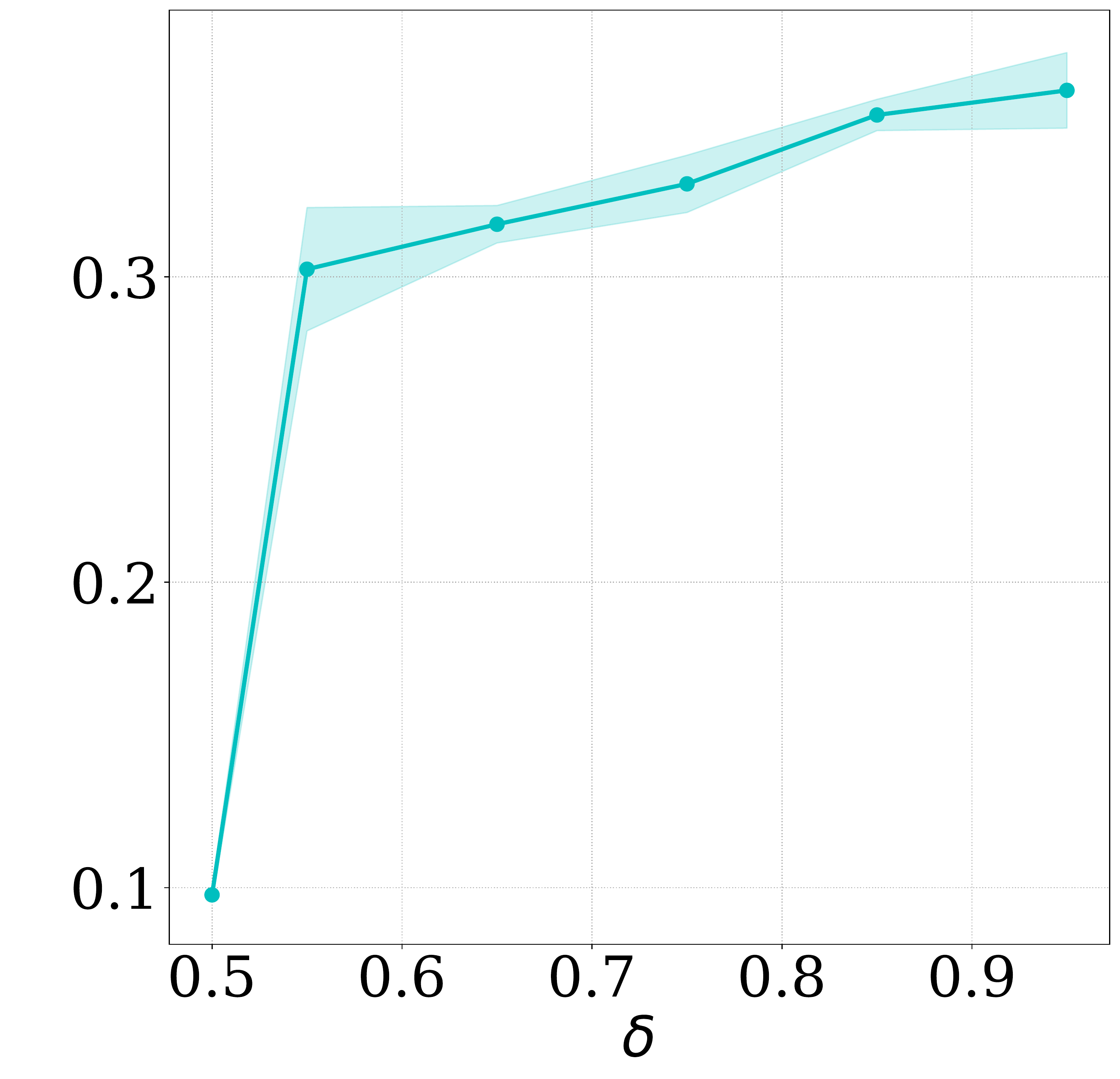}
            \caption{Coverage in Riverswim.}%
            \label{fig:coverage_riverswim_delta}
        \end{subfigure}
        \caption{Coverage in LQG and Riverswim as function of $\delta$; average of 5 seeds, 95\% c.i..}
        \label{fig:deltaTuning}
\end{figure}

In Section~\ref{sec:experiments} we have shown how by tuning $\lambda$ and $\rho$ we can control the amount of exploration. We now show some experiment that illustrate how the hyper parameter $\delta$ can also be use to control exploration, since it defines what percentile of the estimated Gaussian distribution we use as upper bound. The results are reported in Figure~\ref{fig:deltaTuning}. We observe that also $\delta$ is directly related with the coverage. Indeed by increasing $\delta$, we employ larger upper bounds for the value estimates, and this directly translates to larger coverage of the state-action space.

\subsection{Coverage in OAC}

\begin{figure}
        \centering
        \begin{subfigure}[b]{0.49\textwidth}
            \centering
            \includegraphics[width=1\textwidth]{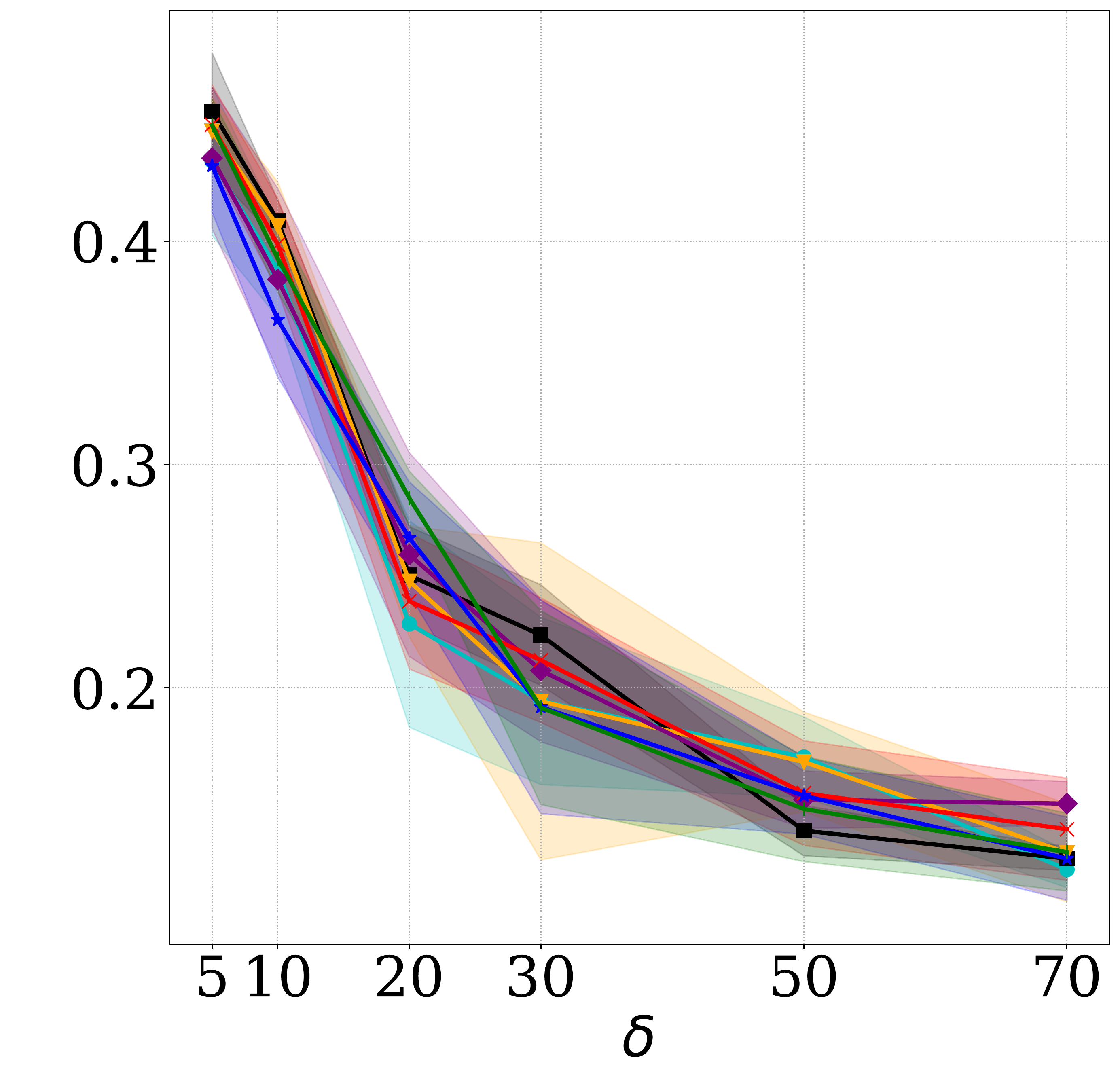}
        \end{subfigure}
        \hfill
        \begin{subfigure}[b]{0.49\textwidth}   
            \centering 
            \includegraphics[width=1\textwidth]{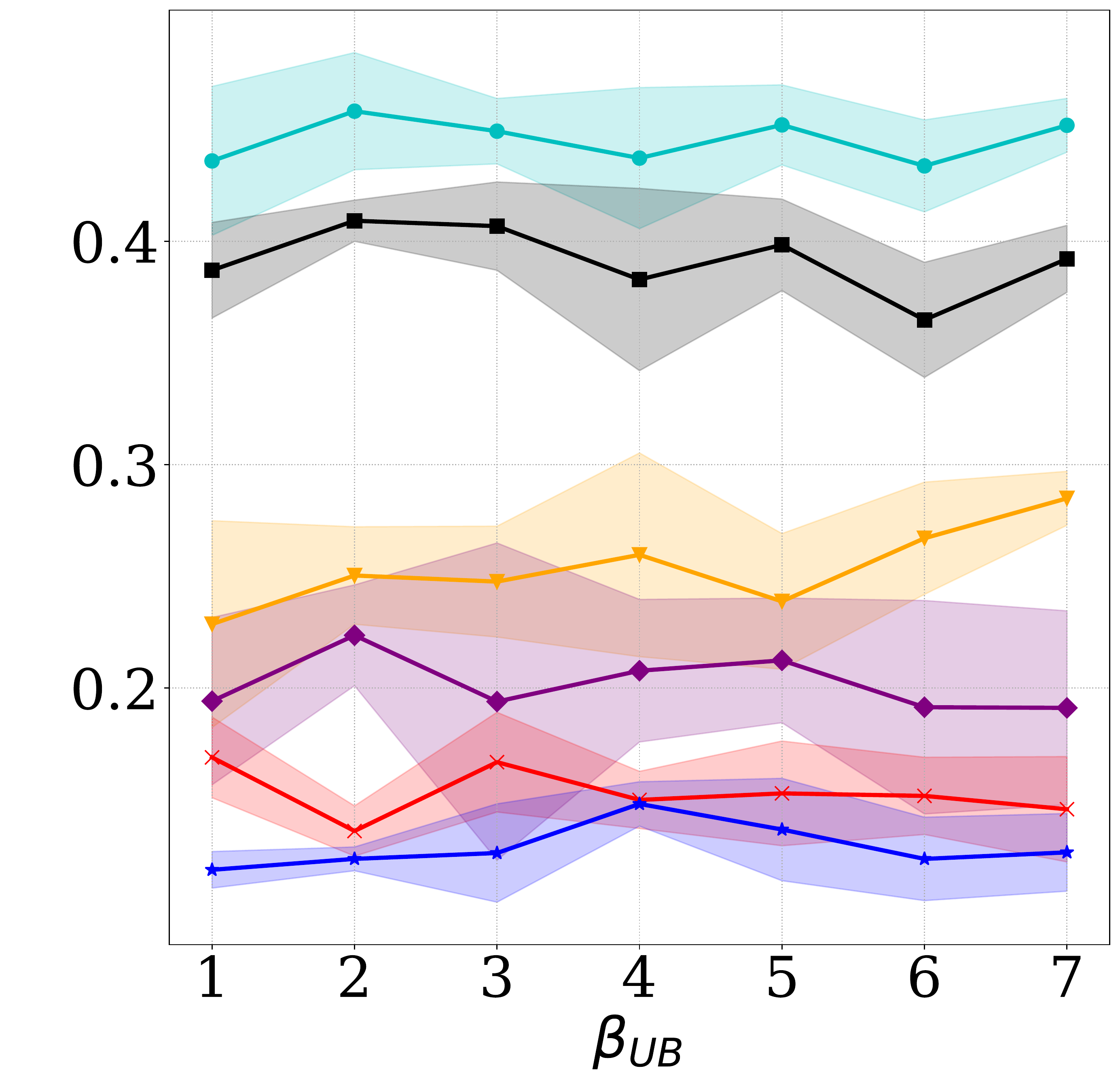}
        \end{subfigure}
        
        \begin{subfigure}[b]{0.49\textwidth}
            \centering
            \includegraphics[width=1\textwidth]{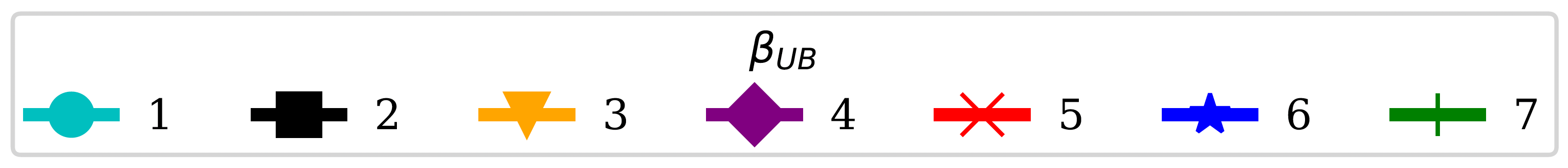}
        \end{subfigure}
        \hfill
        \begin{subfigure}[b]{0.49\textwidth}   
            \centering 
            \includegraphics[width=1\textwidth]{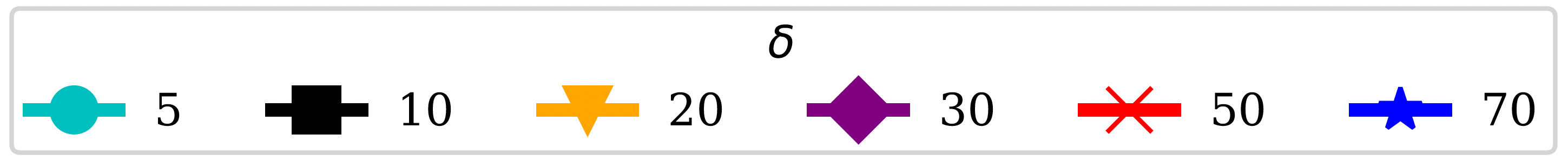}
        \end{subfigure}

        \caption{Coverage in LQG as function of $\delta$ and $\beta_{UB}$; average of 5 seeds, 95\% c.i..}
        \label{fig:OACexpl}
\end{figure}

We performed a similar study to the one presented in Section~\ref{sec:experiments} on WAC for OAC, to investigate whether we could control how much the algorithm explores based on the values of the hyper-parameters. However, what we found is that is hard to predict OAC's exploration based on its hyper parameters $\delta$ and $\beta_{UB}$. In OAC, $\delta$ controls how much the exploration policy differs from the target policy. From Figure~\ref{fig:OACexpl}, we can see that $\delta$ can even negatively affect exploration, if we allow the exploration policy to differ too much from the target policy. The dependence on $\beta$, which controls the definition of the upper bound (similar to our $\delta$ in OAC), suggests that the uncertainty estimate of OAC is not directly related to exploration either. We attribute both these results to the heuristic estimation of uncertainty that OAC employees, based only on the disagreement between the two critics. We argue that this uncertainty estimation is not enough to direct exploration meaningfully.

\subsection{Exploration Heatmaps in Point}

\begin{figure}
        \centering
        \begin{subfigure}[b]{0.49\textwidth}
            \centering
            \includegraphics[width=0.9\textwidth]{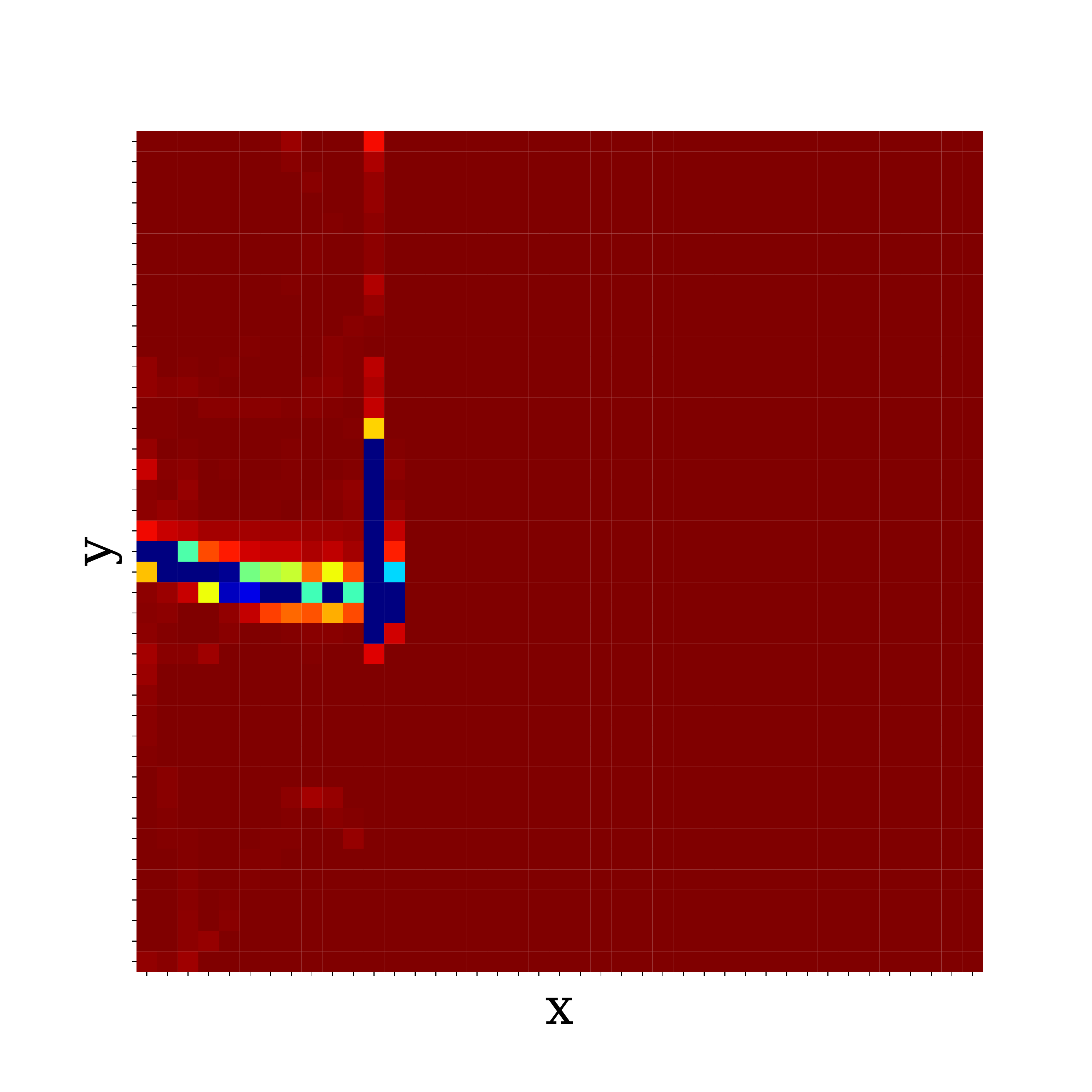}
            \caption{SAC}
            \label{fig:p3d_SAC}
        \end{subfigure}
        \hfill
        \begin{subfigure}[b]{0.49\textwidth}   
            \centering 
           \includegraphics[width=0.9\textwidth]{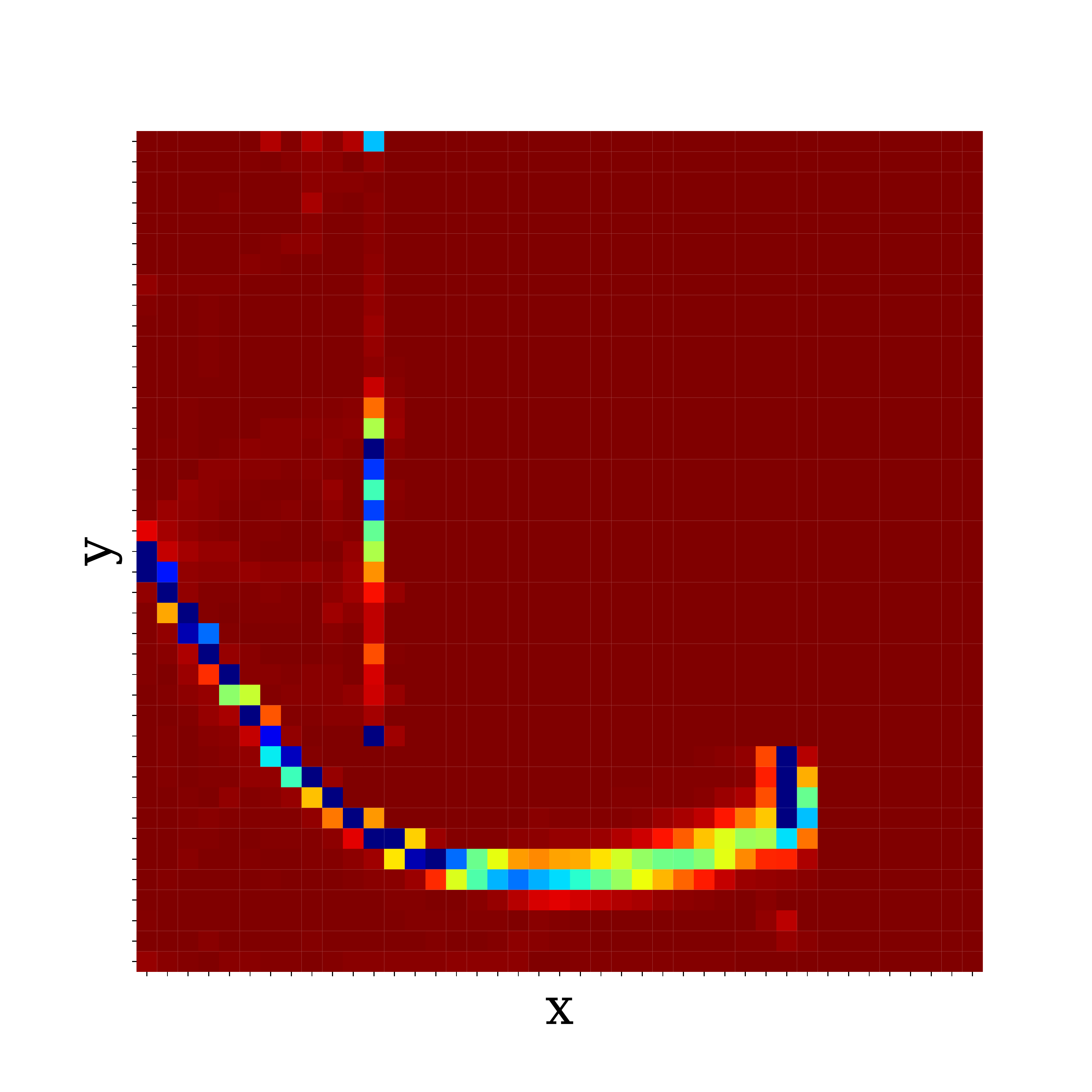}
            \caption{OAC exploration policy}%
            \label{fig:p3d_OAC}
        \end{subfigure}
                \begin{subfigure}[b]{0.49\textwidth}
            \centering
            \includegraphics[width=0.9\textwidth]{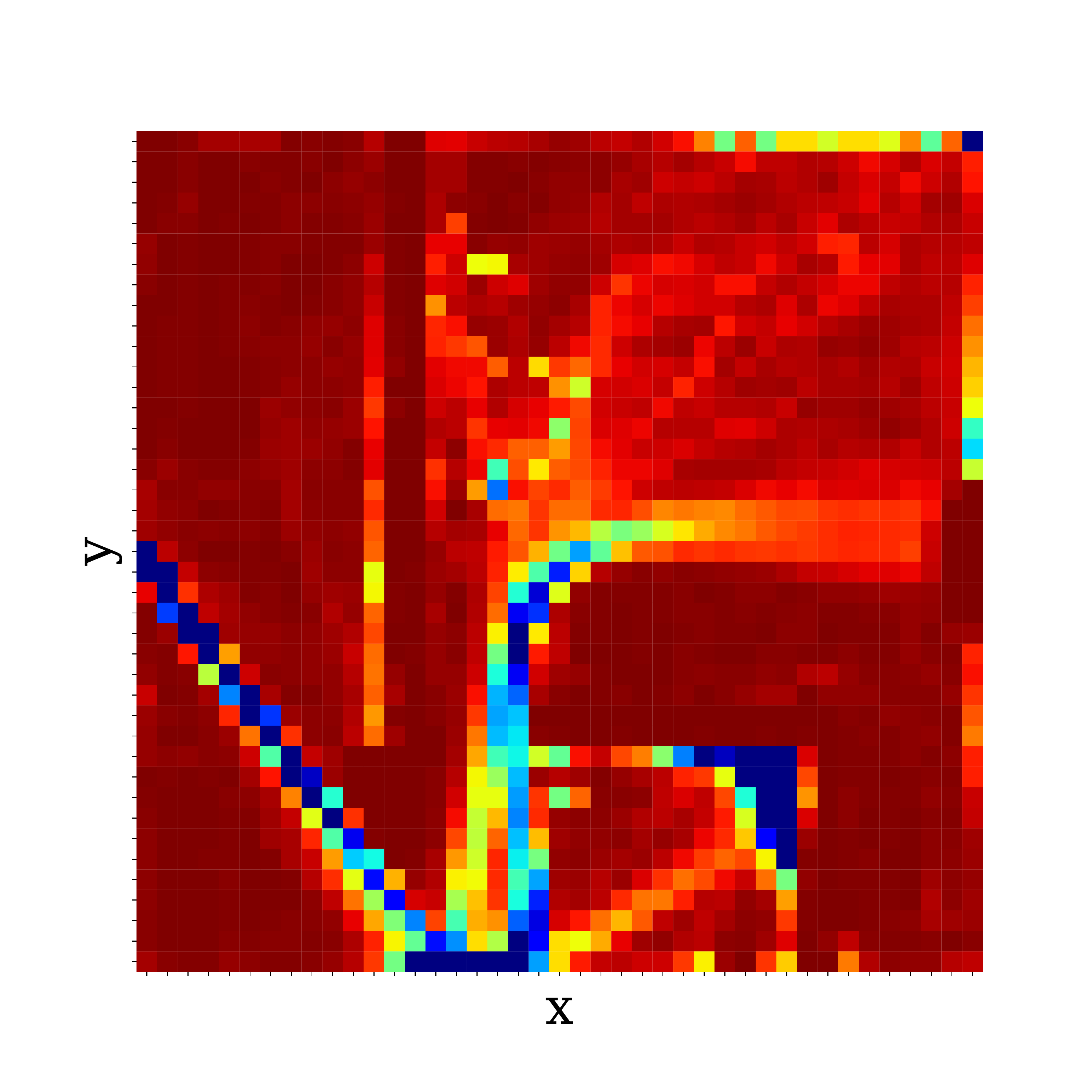}
            \caption{WAC exploration policy}
            \label{fig:p3d_WAC}
        \end{subfigure}
        \hfill
        \begin{subfigure}[b]{0.49\textwidth}   
            \centering 
           \includegraphics[width=0.9\textwidth]{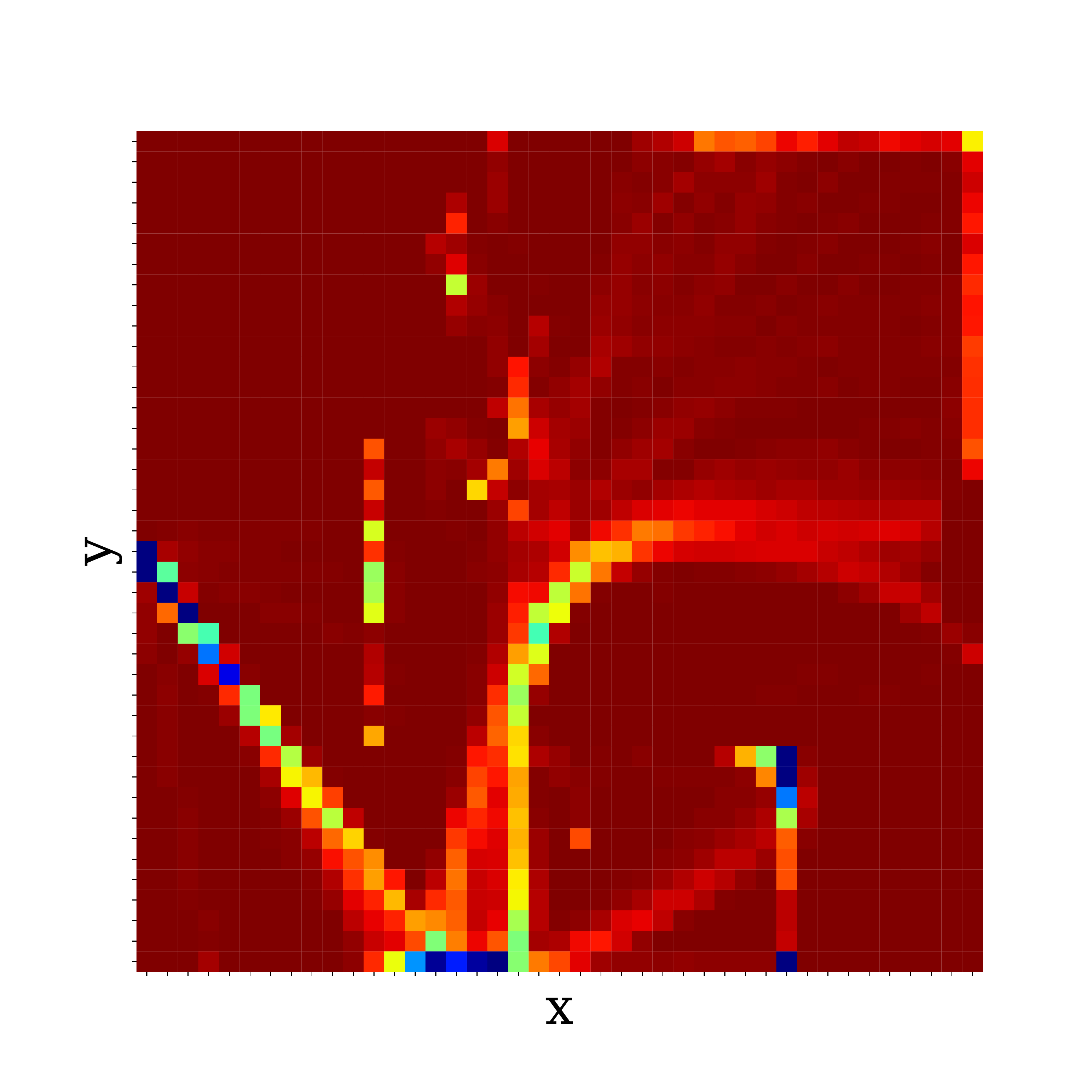}
            \caption{WAC evaluation policy}%
            \label{fig:p3d_WACt}
        \end{subfigure}
        \caption{Cumulative visited states in 300 epochs in Point 3 environment (Dense Reward)}
        \label{fig:point_dense_exploration}
\end{figure}

In this section, we show some additional heatmaps which represents the visited states (we have ignored velocities so we could visualize the location of the agent) over $300$ epochs of all algorithms we have tested throughout the paper. Figure~\ref{fig:point_dense_exploration} shows the heatmaps from runs on Point 3 environment, with dense rewards. In Figure~\ref{fig:p3d_SAC} we see that SAC cannot get past the first wall and does not explore the space around the local maximum enough to reach other maxima. In Figure~\ref{fig:p3d_OAC} we see that OAC finds a better maximum but still a local one. WAC does find the same local maximum as OAC, in fact we can see in Figure~\ref{fig:p3d_WAC} it visits it many times, yet once the uncertainty estimate is low enough it is able to keep exploring and ultimately reach the goal. We have also reported in Figure~\ref{fig:p3d_WACt} the same heatmap created by the target policy which follows the critic of the mean instead of the upper bound.

\begin{figure}
        \centering
        \begin{subfigure}[b]{0.49\textwidth}
            \centering
            \includegraphics[width=0.9\textwidth]{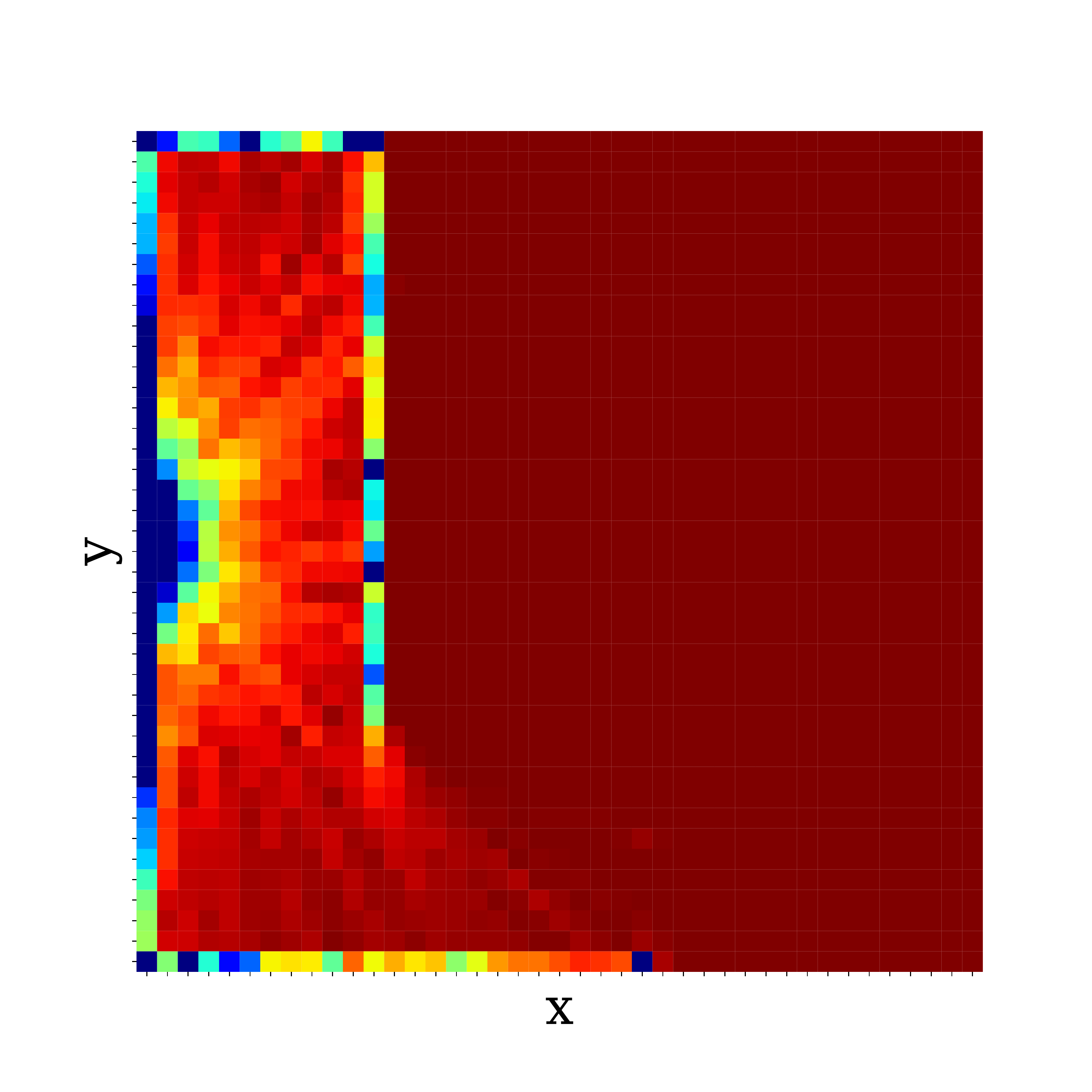}
            \caption{SAC}
            \label{fig:p2s_SAC}
        \end{subfigure}
        \hfill
        \begin{subfigure}[b]{0.49\textwidth}   
            \centering 
           \includegraphics[width=0.9\textwidth]{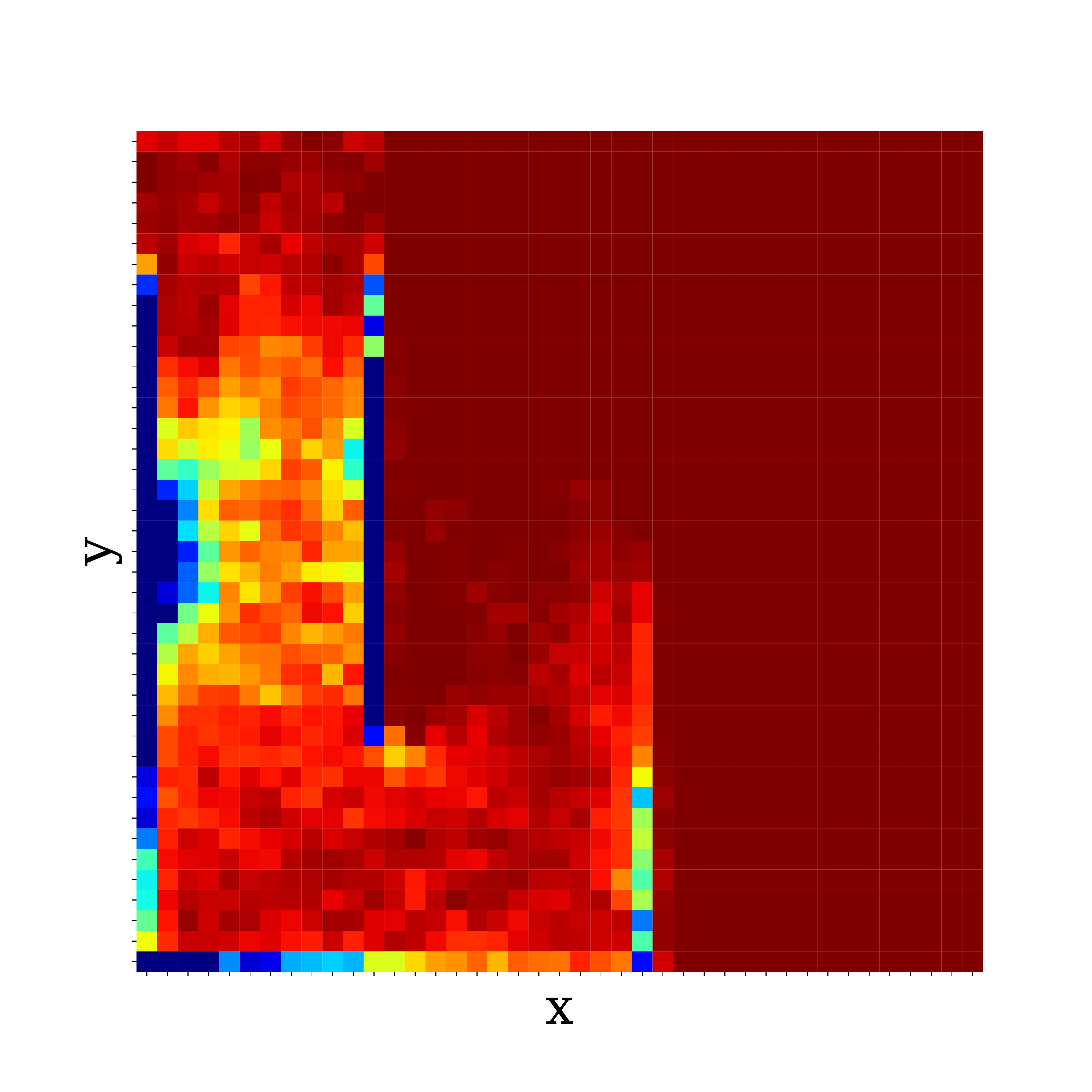}
            \caption{OAC exploration policy}%
            \label{fig:p2s_OAC}
        \end{subfigure}
                \begin{subfigure}[b]{0.49\textwidth}
            \centering
            \includegraphics[width=0.9\textwidth]{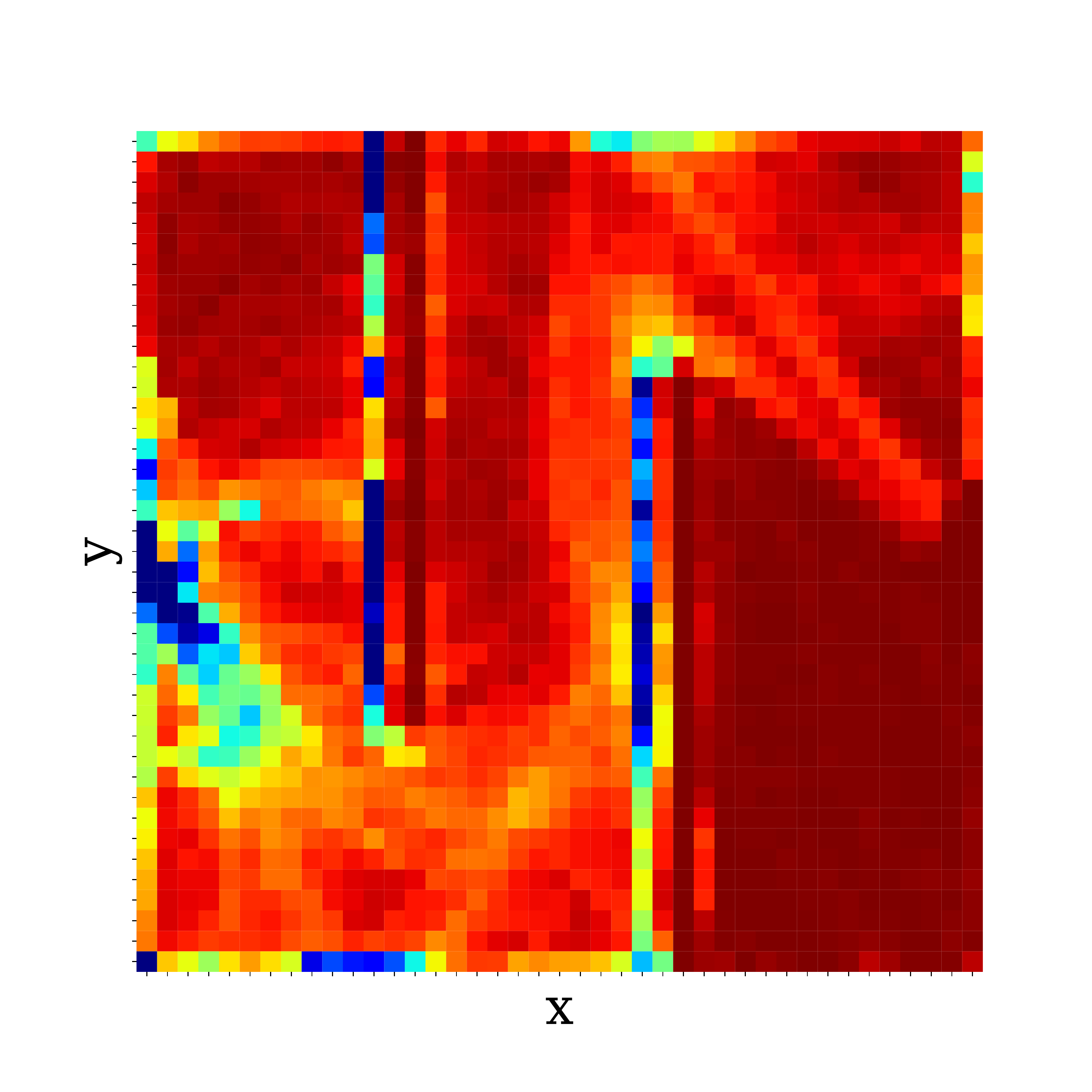}
           \caption{WAC exploration policy}
            \label{fig:p2s_WAC}
        \end{subfigure}
        \hfill
        \begin{subfigure}[b]{0.49\textwidth}   
            \centering 
           \includegraphics[width=0.9\textwidth]{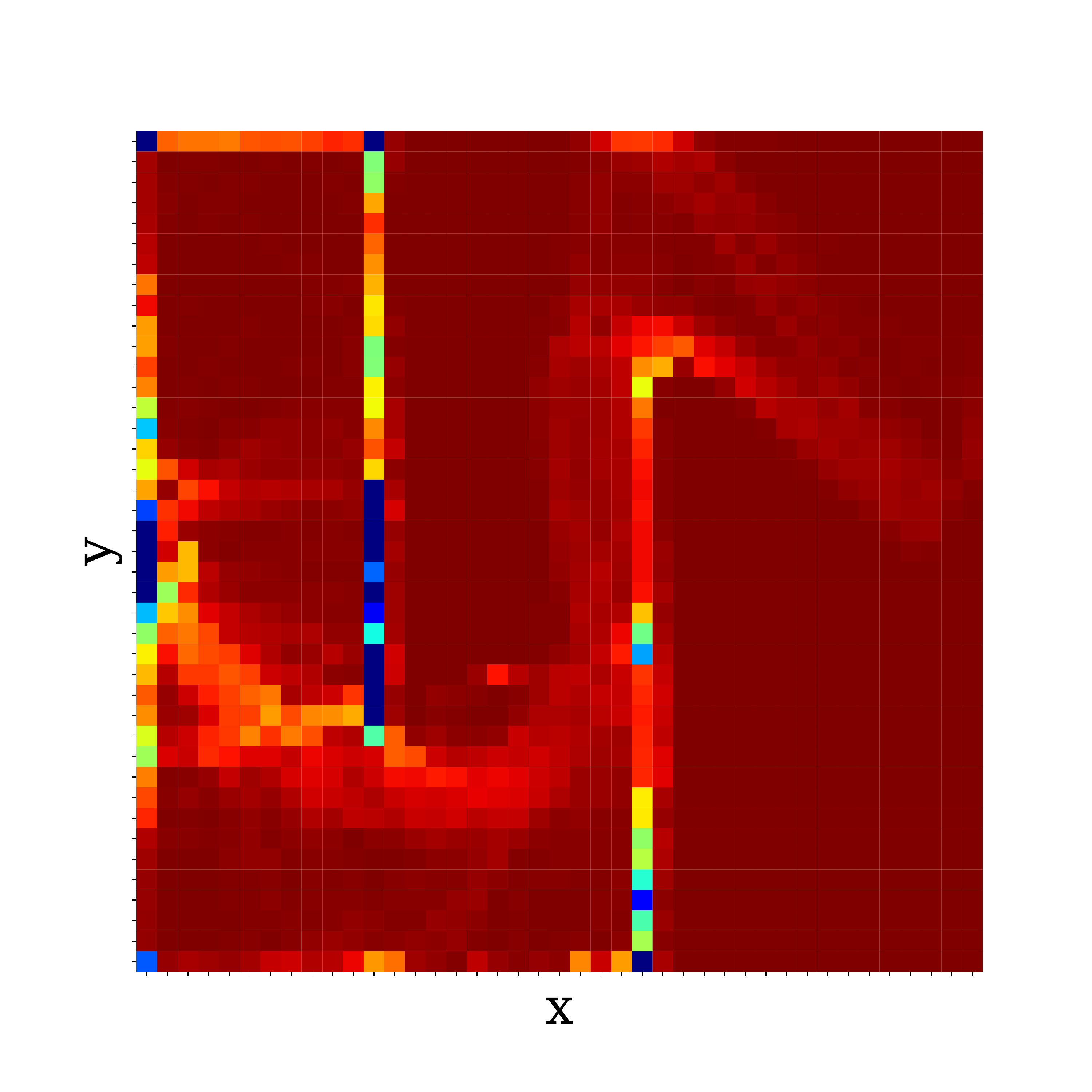}
            \caption{WAC target policy}%
            \label{fig:p2s_WACt}
        \end{subfigure}
        \caption{Cumulative visited states in 300 epochs in Point 2 environment (Sparse Reward)}
        \label{fig:point_sparse_exploration}
\end{figure}

Figure~\ref{fig:point_sparse_exploration} shows the cumulative visited states in the Point 2 environment with sparse reward. We can observe that SAC, having no uncertainty estimate and no informative rewards, mostly explores around the starting states with very simple policies that follow straight lines. OAC is able to reach areas of the maze which are further away from the starting point but it still can't reach the goal. Finally, WAC manages to reach the goal of the maze and explores almost every area of the it. 

\end{document}